\RequirePackage{fix-cm}
\documentclass[twocolumn]{svjour3}          
\smartqed  

\usepackage{graphicx}
\usepackage[numbers,sort&compress]{natbib}
\usepackage{xcolor}
\usepackage{hyperref}
\hypersetup{
 colorlinks=true,
 linkcolor=blue,
 citecolor=blue,
 urlcolor=blue
 }

\usepackage{lineno}
\usepackage{soul}
\usepackage{lscape}
\usepackage{booktabs}
\usepackage{amsmath,mathtools}
\usepackage[intoc]{nomencl}
\usepackage{url}
\usepackage{doi}
\usepackage[boxruled,vlined,linesnumbered]{algorithm2e}
\usepackage{amsfonts}
\usepackage{relsize}
\usepackage{commath}
\usepackage{float}
\usepackage{textcomp}
\usepackage{array}
\usepackage[nottoc,notlot,notlof]{tocbibind}
\usepackage{setspace}
\usepackage{mathptmx}

\usepackage{makecell}
\usepackage{placeins}
\usepackage{dblfloatfix}

\usepackage{subcaption}
\graphicspath{{plots/}}

\usepackage{xcolor}
\definecolor{materialPurple}{HTML}{9C27B0}
\definecolor{materialTeal}{HTML}{009688}
\definecolor{materialGreen}{HTML}{4CAF50}
\definecolor{materialOrange}{HTML}{FF9800}
\definecolor{materialRed}{HTML}{FF0000}
\usepackage{setspace} 
\usepackage{lineno} 
\usepackage{comment}

\usepackage{csquotes}
\usepackage{bm}

\begin{document}

\title{Multi-Asset Closed-Loop Reservoir Management Using Deep Reinforcement Learning
}
\author{Yusuf Nasir \and Louis J.~Durlofsky}

\institute{Yusuf Nasir \at
              Department of Energy Resources Engineering, \\
              Stanford University, Stanford CA 94305, USA \\
              \email{nyusuf@stanford.edu}           
           \and
           Louis J.~Durlofsky \at
              Department of Energy Resources Engineering, \\
              Stanford University, Stanford CA 94305, USA \\
              \email{lou@stanford.edu}
}

\date{Received: date / Accepted: date}
\maketitle

\begin{abstract} Closed-loop reservoir management (CLRM), in which history matching and production optimization are performed multiple times over the life of an asset, can provide significant improvement in the specified objective. These procedures are computationally expensive due to the large number of flow simulations required for data assimilation and optimization. Existing CLRM procedures are applied asset by asset, without utilizing information that could be useful over a range assets. Here, we develop a CLRM framework for multiple assets with varying numbers of wells. We use deep reinforcement learning to train a single global control policy that is applicable for all assets considered. The new framework is an extension of a recently introduced control policy methodology for individual assets. Embedding layers are incorporated into the representation to handle the different numbers of decision variables that arise for the different assets. Because the global control policy learns a unified representation of useful features from multiple assets, it is less expensive to construct than asset-by-asset training (we observe about $3\times$ speedup in our examples). The production optimization problem includes a relative-change constraint on the well settings, which renders the results suitable for practical use. We apply the multi-asset CLRM framework to 2D and 3D water-flooding examples. In both cases, four assets with different well counts, well configurations, and geostatistical descriptions are considered. Numerical experiments demonstrate that the global control policy provides objective function values, for both the 2D and 3D cases, that are nearly identical to those from control policies trained individually for each asset. This promising finding suggests that multi-asset CLRM may indeed represent a viable practical strategy.



\end{abstract}

\section{Introduction}
\label{sec:1}

The traditional closed-loop reservoir management (CLRM) framework involves the continuous collection of well data and the repeated application of history matching followed by production optimization over the set of updated geological models. In the robust production optimization step, well settings such as time-varying injection/production rates or bottomhole pressures (BHPs) are determined. CLRM thus provides uncertainty reduction and value maximization for a prescribed objective function, often taken to be net present value, NPV, of the project. The  history matching and optimization problems that must be repeatedly solved in the CLRM workflow are, however, computationally expensive, and this limits the use of CLRM in practical applications. An important limitation of existing CLRM procedures is that they are applicable on a per-asset basis, i.e., they do not leverage information from one system to another. Procedures that can learn, and thus utilize information across multiple assets, could lead to significantly reduced computational requirements, thus enabling the wider use of the overall CLRM methodology.

In recent work~\citep{nasir2022deep}, we developed a control-policy approach based on deep reinforcement learning (DRL) for efficient CLRM. The control policy, which is represented by a deep neural network that is trained in a preprocessing (offline) step, maps observed production and injection data to optimal well pressure settings. These treatments, however, are applicable only for a single asset with a fixed number of existing wells. 
Our goal in this work is to extend this framework to enable the training of a single control policy that can prescribe optimal well settings for multiple assets with different numbers of wells in different locations. To achieve this multi-asset CLRM, we introduce treatments that allow the global control policy to account for a varying number of decision variables across the different assets. The training of the global policy requires much less computational cost than would be expended in training a control policy for each asset independently.

Traditional CLRM procedures have been presented by a number of investigators, e.g., \citep{brouwer2004improved, aitokhuehi2005optimizing, sarma2006efficient, jansen2009closed}. The robust optimization step in these frameworks is achieved through use of derivative-free~\citep{onwunalu2010application, bouzarkouna2012well, isebor2014a, nasir2020hybrid} or gradient-based~\citep{wang2002optimization, alhuthali2008optimal, liu2020sequential} optimizers. History matching is accomplished using optimization-based \citep{dadashpour2010derivative, liu2020multilevel} or ensemble-based \citep{chen2006data, emerick2013ensemble} procedures. None of the traditional CLRM implementations is for multiple assets. These approaches also tend to find overly conservative solutions due to the assumption of models being equally probable in the robust optimization steps. This assumption can have an especially large impact when nonlinear output constrains are prescribed, because these constraints must be satisfied in all (or nearly all) models considered in the robust optimization. 

The use of control policies~\citep{addiego2008insurance, dilib2015closed, hanssen2017closed, nasir2022deep} has proven to be an effective technique to address the conservative nature of robust optimization solutions in reservoir management problems. These control policies are designed to map observed data to optimal well settings. As such, they use the production history in determining future actions (well settings). The DRL-based general control-policy procedure introduced in~\citep{nasir2022deep} was shown to provide better objective-function values than the traditional CLRM framework (which is based on robust optimization). DRL has also been applied to provide optimal well settings in deterministic systems, where the reservoir model is assumed to be known. For example, Ma et al.~\citep{ma2019waterflooding}, Miftakhov et al.~\citep{miftakhov2020deep}, and Zhang et al.~\citep{zhang2022training} trained control policies that map quantities such as the pressure and saturation distribution to optimal well settings. In previous applications of DRL for the well control optimization problem, in both  deterministic~\citep{ma2019waterflooding, miftakhov2020deep, zhang2022training} and uncertain~\citep{nasir2022deep} systems, a single asset with a given number of existing wells was considered. 

Multitask learning~\citep{zhang2021survey}, where a deep neural network is trained to efficiently solve multiple tasks, has been explored in the context of both supervised learning~\citep{maurer2016benefit, long2017learning} and reinforcement learning~\citep{brunskill2013sample, cheng2022provable}. Cheng et al.~\citep{cheng2022provable} showed that learning a global representation of the features from multiple tasks is more efficient than learning feature representations for each task individually. Other investigators have considered the use of evolutionary multitask optimization~\citep{osaba2022evolutionary} techniques to solve different tasks simultaneously. In contrast to multitask learning, evolutionary multitask optimization does not involve learning a global representation of features from the tasks. Rather, knowledge transfer between the tasks occurs in the space of decision variables through operations that are usually heuristically defined. As noted in~\citep{osaba2022evolutionary}, the efficiency of any procedure in solving multiple optimization tasks largely depends on its ability to learn and exploit useful global information among tasks. Yao et al.~\citep{yao2021self} applied a multifactorial evolutionary algorithm for production optimization involving reservoirs with different rock and fluid properties, but similar well patterns. The proposed approach in~\citep{yao2021self} was applied to assets where the reservoir models are assumed to be known, i.e., one reservoir model per asset. Our goal in this work, however, is to find a global policy that provides decisions based on observed data for multiple uncertain assets.

He et al.~\citep{he2021deep} and Nasir et al.~\citep{nasir2021deep} proposed the use of DRL for generalizable field development optimization for single-phase~\citep{he2021deep} and two-phase~\citep{nasir2021deep} flow problems (field development involves the placement of new wells). Their procedure determines the well count, location and drilling schedule for different reservoirs under varying economic conditions. These reservoirs can vary in terms of fluid and rock properties. In~\citep{he2021deep, nasir2021deep}, a fixed-size reservoir template was used for the policy training. Thus the dimension of the discrete well-location decision variables was the same for all assets. An action-masking technique~\citep{huang2020closer} was used to avoid the consideration of invalid drilling locations. This masking technique entails setting the probability of sampling invalid drilling locations to zero. While this approach is efficient for eliminating invalid discrete decision variables, it is not applicable to the multi-asset CLRM problem because this case involves varying numbers of continuous decision variables.

In this paper, we extend the DRL-based general control policy formulation introduced in \citep{nasir2022deep} to the multi-asset CLRM setting. New treatments include the use of an embedding layer in the control policy architecture to enable the output of varying numbers of continuous decision variables, which are represented by Gaussian distributions.
During the policy training, well settings for the different assets are sampled from these distributions. We also incorporate treatments to avoid impractical changes in the well settings (e.g., large jumps) from one control step to another. The new framework is applied to train control policies for 2D and 3D water-flooding problems. In the 2D case, we consider assets of the same areal dimension, while in the 3D case the reservoir models vary in both areal dimension and in the number of layers. Results from the new multi-asset CLRM framework are compared to those obtained from control policies established for each asset individually.

This paper proceeds as follows. In Section~\ref{sec:cp_clrm}, we provide a brief description of the general control policy formulation introduced in~\citep{nasir2022deep} for single-asset CLRM. This includes the single-asset control policy representation and the training procedure. The extension of these treatments to the multi-asset case is presented in Section~\ref{sec:ma_clrm}. Numerical experiments comparing the multi-asset and single-asset CLRM formulations for assets represented by 2D and 3D geological models are provided in Section~\ref{sec:results}. We conclude in Section~\ref{sec:conclusion}
with a summary and suggestions for future work. 

\section{DRL for control-policy-based single-asset CLRM}
\label{sec:cp_clrm}

In this section, we briefly describe the DRL-based control-policy procedure for CLRM for a single asset~\citep{nasir2022deep}. We introduce a treatment for the parameterization of well settings, which facilitates the application of constraints that are useful in practice. We then discuss control policy optimization and the neural network representation for the single-asset case.

\subsection{Control-policy-based CLRM}

The goal in single-asset CLRM is to determine the well settings -- taken to be bottomhole pressures (BHPs) in this work -- for $N_w$ wells at different decision-making/control steps. The general control-policy approach for the single-asset CLRM entails solving an offline optimization problem given by
\begin{gather}
\begin{array}{rrclcl}
\displaystyle \max_{\boldsymbol{\theta} \in \mathbb{R}^{N}} & {J (\pi_{\theta}, M)}, \ \  \textrm{subject to: } \textbf{c}(\pi_{\theta})\leq\textbf{0}, \\
& |\frac{u_k^w - u_{k-1}^w}{u_{k-1}^w}|\leq \epsilon_c\epsilon_{max}^w,
\end{array}
\label{eq:policy_opt}
\end{gather}
where $\pi_{\theta}$ is the control policy, defined by parameters $\boldsymbol{\theta} \in \mathbb{R}^N$, which maps observed data to well settings, and $M$ is a set of $N_r$ prior geological models, with $M = {\{\textbf{m}^{1}, \textbf{m}^{2}, \ldots, \textbf{m}^{N_{r}}\}}$. The prior realizations $M$ can be samples from multiple geological scenarios. The vector \textbf{c} denotes nonlinear output constraints such as the maximum well liquid production rate or water injection rate. The well setting for well $w$ at control step $k$, $u_k^w$, is constrained to be within a specified range based on the well setting at the previous control step, $u_{k-1}^w$. The limit on the relative change of the well setting is governed by the parameters $\epsilon_{max}^w$ and $\epsilon_c$. Here $\epsilon_{max}^w = (u_{ub}^w-u_{lb}^w)/{u_{lb}^w}$ is the maximum allowable relative change, which would shift the well setting from its lower bound $u_{lb}^w$ to its upper bound $u_{ub}^w$, 
and $\epsilon_c \in [0, 1]$ is a user-defined parameter that controls the allowable relative change. Specifying $\epsilon_c = 1$ removes the constraint on the changes in well settings, while $\epsilon_c = 0$ requires the specification of a single well setting, for each well, throughout the production life. 


The objective function in Eq.~\ref{eq:policy_opt}, $J(\pi_{\theta}, M)$, is defined as the average net present value (NPV) over the $N_r$ realizations. This is computed as
\begin{gather}
	J(\pi_{\theta}, M) = \frac{1}{N_r} \sum\limits_{i = 1}^{N_r} \sum\limits_{k = 1}^{N_c} \textnormal{NPV}(\textbf{u}_{k}^{i}, \ \textbf{m}^{i}),
    \label{eq:policy_exp_npv}
\end{gather} 
where $N_c$ is the number of control steps. The vector of well settings for geological model $\textbf{m}^{i}$ in control step $k$ is defined as $\textbf{u}_{k}^{i} = \pi_{\theta}(\textbf{u}_{k-1}^{i}, \epsilon_c, D_{1:k}^{i})$, with $D_{1:k}^{i} = \{D_1^i, D_2^i, \ldots, D_k^i\}$ the set of all observed data, for geological model $\textbf{m}^{i}$, from the first control step to the beginning of control step $k$. The observed data $D_k \in \mathbb{R}^{N_{d}\times(3N_p + 2N_i)}_+$ comprise the oil production rate, water injection rate, watercut and BHPs constrained to satisfy the nonlinear constraints $\textbf{c}(\pi_{\theta})$. These quantities are reported at $N_d$ regular intervals at each control step for $N_p$ production wells and $N_i$ injection wells.

The term $\textnormal{NPV}(\textbf{u}_{k}^{i}, \ \textbf{m}^{i})$ is the contribution at control step $k$ to the NPV for geological model $\textbf{m}^{i}$. This contribution, which only involves simulation data at step $k$, is computed as
\begin{equation}
\begin{aligned}
& \textnormal{NPV}(\textbf{u}_{k}^{i}, \ \textbf{m}^{i}) = \\
& \sum\limits_{j = 1}^{N_t}  \frac{\left[\sum\limits_{w = 1}^{N_p} \left(p_{o} ~q^{w}_{o,j}-c_{pw}~q^{w}_{pw,j}\right) - \sum\limits_{w = 1}^{N_i} c_{iw}~q^{w}_{iw,j}\right] - c_{opex}}{( 1+b )^{t_j/365}}\Delta t_j,
\end{aligned}
\label{eq_npv}
\end{equation}
where $c_{opex}$, $p_{o}$, $c_{pw}$, $c_{iw}$ and $b$ represent the operating cost per day, oil price, cost of produced and injected water, and the annual discount rate, respectively. The variables $N_t$, $N_i$ and $N_p$ denote the number of time steps, number of injectors, and number of producers. The time and time step size (in days) at time step $j$ are denoted as $t_j$ and $\Delta t_j$. The quantities $q^{w}_{o,j}$, $q^{w}_{pw,j}$, and $q^{w}_{iw,j}$ are the rates of oil and water production and water injection, for well $w$ at time step $j$. Production operations are terminated at any control step where $\textnormal{NPV}(\textbf{u}_{k}^{i}, \ \textbf{m}^{i}) < 0$, which can occur when the field watercut approaches 1. This treatment ensures that all control steps have a positive cash flow.

\subsection{Control policy optimization procedure}

\begin{figure}[htp]
	\centering
	\includegraphics[width=8cm]{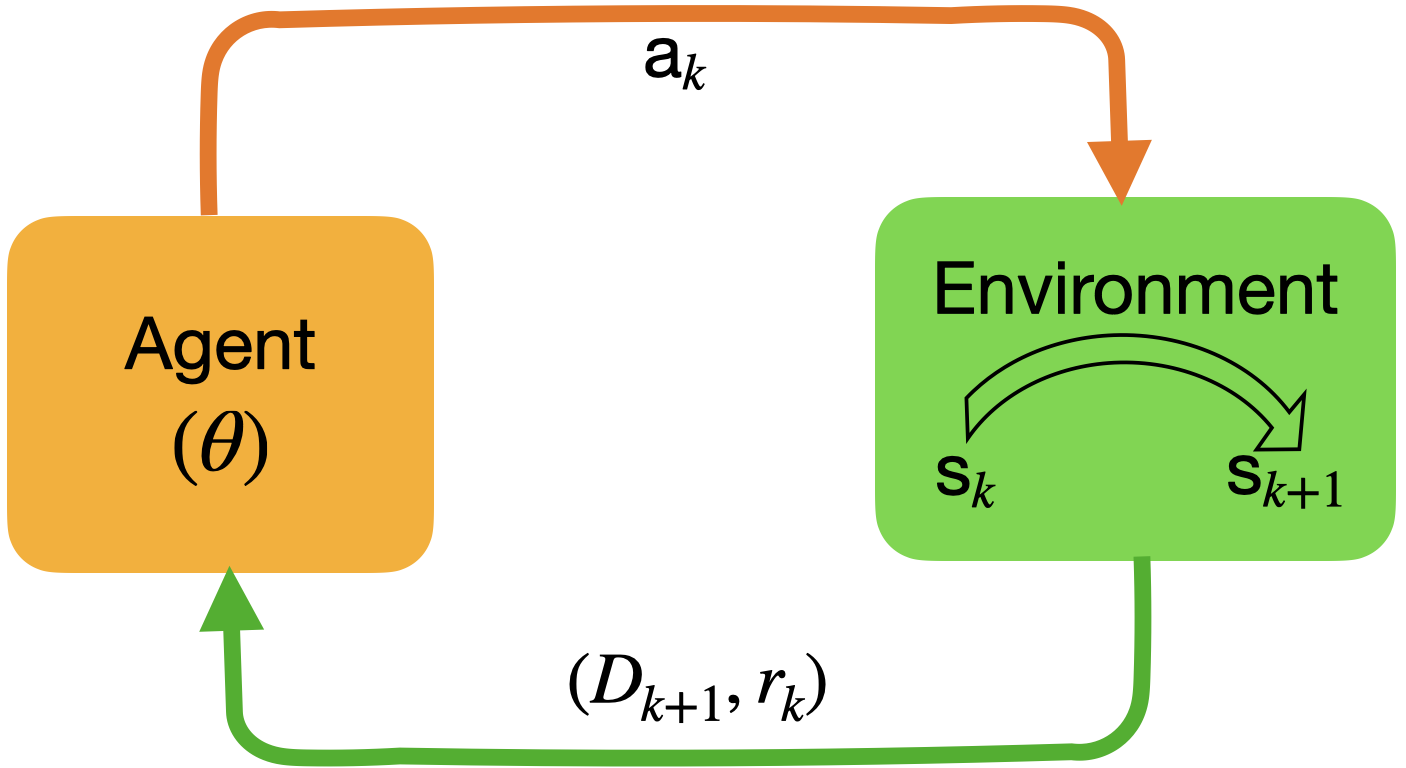}
	\caption{Deep reinforcement learning agent-environment interface. Figure modified from~\citep{nasir2022deep}.}
	\label{fig:drl_interface}
\end{figure}

The optimization problem posed in Eq~\ref{eq:policy_opt} is solved using DRL. In DRL, the control policy (also referred to as the agent) is represented by a deep neural network that interacts with an environment, which in our case is the reservoir model. The setup is illustrated in Fig.~\ref{fig:drl_interface}. The interaction between the control policy and the environment is governed by the decisions/actions $\textbf{a}_k$ taken by the policy in each control step, and the resulting observation $D_{k+1}$ and reward $r_k$ provided by the environment. The goal of the policy is to select actions based on all observed data in order to maximize the cumulative reward. The CLRM problem is formulated as a partially observable Markov decision process (POMDP), where the state quantities $\textbf{s}_k$ (such as the pressure and saturation distribution) that define the transition of the environment from control step to control step are unobserved. 

The actions $\textbf{a}_k \in [-1, 1]^{N_w}$ define the relative change in the well settings at control step $k$. The BHP for well $w$, $u_k^w$, is given by  
\begin{subequations}
\begin{align}
\label{eq:def_well_settings}
\begin{split}
\tilde{u}_{k}^w = u_{k-1}^w(1 + a_{k}^w\epsilon_c\epsilon_{max}^w),
\end{split}
\\[2ex]
\begin{split}
u_{k}^w = clip(\tilde{u}_{k}^w, u_{lb}^w, u_{ub}^w),
\end{split}
\end{align}
\end{subequations}
where $clip$ ensures the BHP is between the lower and upper bounds. The well settings for the first control step are obtained by linearly mapping $\textbf{a}_1$ within the BHP bounds.



The control policy training procedure entails the determination of the policy parameters $\boldsymbol{\theta}$ that maximize the cumulative reward, with the reward for each control step $r_k^i$ defined as $r_k^i = \textnormal{NPV}(\textbf{u}_k^i, \ \textbf{m}^i)$. During training, realizations are sampled from $M$, $\epsilon_c$ is sampled from [0, 1], and actions are sampled from a Gaussian distribution (defined by the control policy) that represents the continuous space $[-1, 1]^{N_w}$. 
Policy optimization is performed using the proximal policy optimization (PPO)~\citep{schulman2017proximal} algorithm. We note that PPO has been successfully applied to field development~\citep{he2021deep, nasir2021deep} and reservoir management~\citep{miftakhov2020deep, nasir2022deep} problems.

\subsection{Control policy representation}

\begin{figure*}[htbp!]
	\centering
	\includegraphics[width=0.8\textwidth]{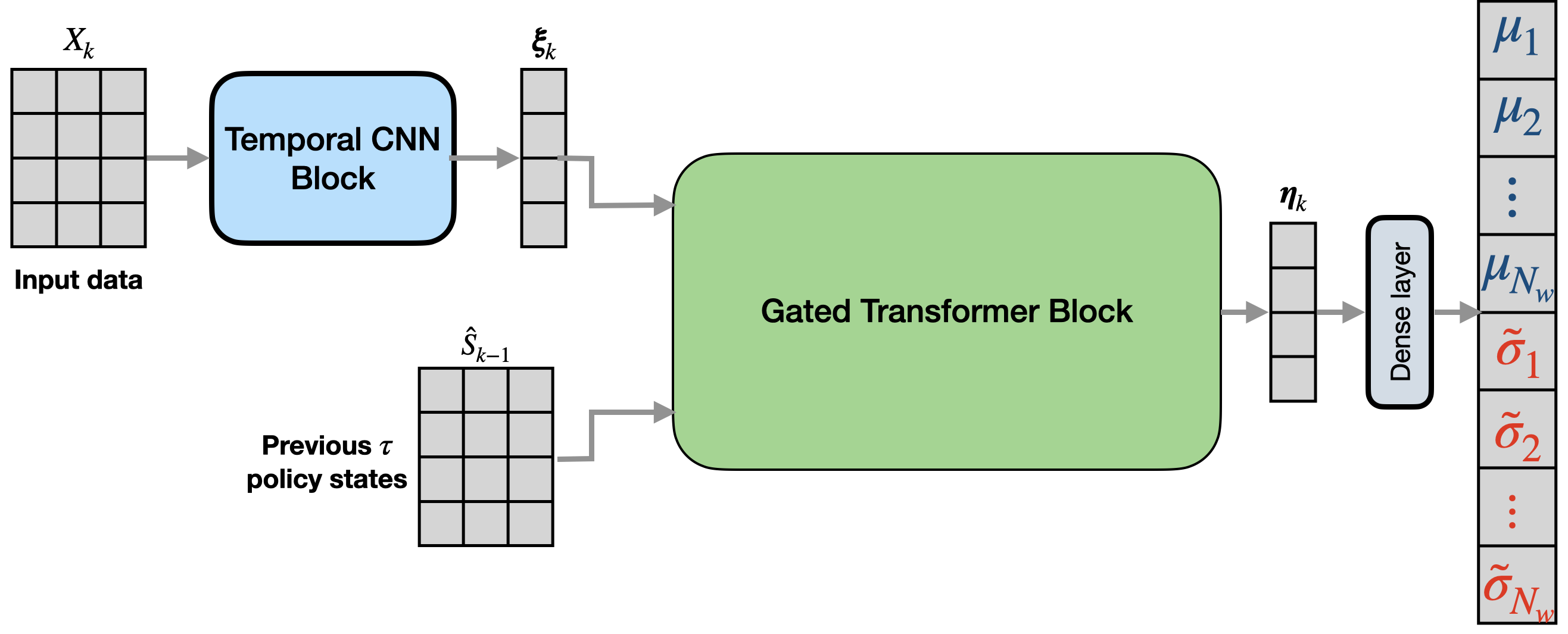}
	\caption{Single-asset control policy represented by temporal and gated transformer blocks. The input $X_k$ contains current observed data, well settings from the previous control step, and the sampled allowable relative change in well settings. The output is the action distribution defined by the action mean and log-standard deviation.}
	\label{fig:policy_single_asset}
\end{figure*}

The control policy $\pi_{\theta}$ for the single-asset case is represented by the memory-based neural network shown in Fig.~\ref{fig:policy_single_asset}. This network enables a concise representation of all acquired data, up to control step $k$, through a policy state $\hat{\textbf{s}}_{k}$ that is updated via
\begin{equation}
   \hat{\textbf{s}}_{k} = f_{\zeta}(\hat{S}_{k-1}, \textbf{u}_{k-1}, \epsilon_c, D_{k}),
    \label{eq:mem_upd}
\end{equation}
where $f_\zeta$ denotes the policy state update function, with $\boldsymbol{\zeta} \subset \boldsymbol{\theta}$, and $\hat{S}_{k-1} = \{\hat{\textbf{s}}_{k-1}, \hat{\textbf{s}}_{k-2}, \ldots, \hat{\textbf{s}}_{k-\tau}\}$. Here $\tau$ is a user-specified number of previous policy states used in computing the current policy state. In this work, based on numerical experimentation, we specify $\tau=5$.

The input data to the policy is represented by a matrix $X_k \in \mathbb{R}^{N_{d}\times(4N_p + 3N_i + 1)}_+$. This matrix is comprised of the observations $D_{k} \in \mathbb{R}^{N_{d}\times(3N_p + 2N_i)}_+$, the well settings vector $\textbf{u}_{k-1}$ repeated $N_d$ times into a matrix of dimension $N_{d}\times (N_p + N_i)$, and the scalar $\epsilon_c$ repeated $N_d$ times into a vector of dimension $N_d \times 1$. 
This treatment allows us to have an input that can be processed by the 1D convolutional neural network (CNN) layers in the temporal CNN block in Fig.~\ref{fig:policy_single_asset}. The feature vector $\boldsymbol{\xi}_k \in \mathbb{R}^{N_m}$ is the output from the temporal CNN block after processing $X_k$. 

The previous policy states $\hat{S}_{k-1}$ and $\boldsymbol{\xi}_k$ serve as input to a gated transformer block~\citep{parisotto2020stabilizing}. This block is composed of relative multihead attention (RMHA) and multilayer perceptron (MLP) submodules. The RMHA module extracts the relevant information from the inputs $\hat{S}_{k-1}$ and $\boldsymbol{\xi}_k$ through an attention operation~\citep{vaswani2017attention}. Output features from the RMHA module are further processed by dense (fully connected) layers in the MLP module. The final output from the gated transformer block $\boldsymbol{\eta}_k \in \mathbb{R}^{N_m}$ is processed by a dense layer to obtain the action distribution of the policy defined by the action mean $\boldsymbol{\mu}_k \in \mathbb{R}^{N_w}$ and action log-standard deviation $\tilde{\boldsymbol{\sigma}}_k \in \mathbb{R}^{N_w}$. The output from the neural network is taken as the log-standard deviation because it can be negative. 
Actions $\textbf{a}_k$ are sampled from the Gaussian distribution defined by the mean $\mu_k^w$ and standard deviation $\sigma_k^w = \exp{(\tilde{\sigma}_k^w)}$. Please refer to~\citep{nasir2022deep} for further details on the policy representation for the single-asset CLRM. 

\section{DRL for control-policy-based multi-asset CLRM}
\label{sec:ma_clrm}

The DRL-based control policy approach described in Section~\ref{sec:cp_clrm} is applicable for a single asset. The asset was assumed to have a fixed number of wells $N_w$, and the set of prior realizations $M$ was conditioned to the hard data at the $N_w$ well locations. In this section we extend this approach to construct a single global control policy applicable to multiple assets with varying numbers of wells in different locations. 

\subsection{Control-policy-based multi-asset CLRM}

In the multi-asset CLRM, the goal is to find a global policy $\pi_{\theta_g}$, defined by parameters $\boldsymbol{\theta}_g \in \mathbb{R}^{N_g}$, for $N_a$ different assets. The global control policy $\pi_{\theta_g}$ learns a global representation of the flow behavior for the $N_a$ assets through observed data. This global representation is encoded in the shared parameters $\boldsymbol{\theta}_g$. As will be shown later, this approach is more computationally efficient than training $N_a$ individual control policies for the different assets.

The global control policy optimization is
\begin{gather}
\begin{array}{rrclcl}
\displaystyle \max_{\boldsymbol{\theta}_g \in \mathbb{R}^{N_g}} & {J (\pi_{\theta_g}, M_g)}, \ \  \textrm{subject to} \  \textbf{c}(\pi_{\theta_g})\leq\textbf{0}, \\
& |\frac{u_k^{n,w} - u_{k-1}^{n,w}}{u_{k-1}^{n,w}}|\leq \epsilon_c\epsilon_{max}^{n,w},
\end{array}
\label{eq:ma_policy_opt}
\end{gather}
where $u_k^{n,w}$ is the well setting for well $w$ in asset $n$, for $n=1, \dots N_a$, in control step $k$. We consider an equal number of prior realizations, $N_r$, for each asset. The set of $N_a \times N_r$ prior realizations that enter the global policy optimization is denoted by $M_g = [M^1 \ M^2 \hdots M^{N_a}]$, where $M^{n}$, $n=1, \dots N_a$, is the set of prior realizations for asset $n$. The number of wells in asset $n$ is denoted $N_w^n$, and the prior realizations $M^n$ are conditioned to hard data at the $N_w^n$ well locations. The prior realizations for each asset can be sampled from a range of geological scenarios. The geometries and grid dimensions of the geological models for the $N_a$ assets can differ, as this information does not enter the policy representation.
 
The objective function $J(\pi_{\theta_g}, M_g)$ for the global policy optimization is given by
\begin{gather}
	J(\pi_{\theta_g}, M_g) = \frac{1}{N_aN_r} \sum\limits_{n = 1}^{N_a}\sum\limits_{i = 1}^{N_r} \sum\limits_{k = 1}^{N_c} \textnormal{NPV}(\textbf{u}_{k}^{n,i}, \ \textbf{m}^{n,i}),
    \label{eq:ma_policy_exp_npv}
\end{gather} 
where the vector $\textbf{u}_k^{n,i} \in \mathbb{R}_+^{N_w^n}$ defines the well settings for geological model $\textbf{m}^{n,i}$ in control step $k$, and $\textbf{m}^{n,i}$ denotes realization $i$ of asset $n$. The definition of $\textnormal{NPV}(\textbf{u}_{k}^{n,i}, \ \textbf{m}^{n,i})$ is similar to Eq.~\ref{eq_npv}, with the number of production and injection wells for asset $n$ given as $N_p^n$ and $N_i^n$. The well settings vector $\textbf{u}_k^{n,i}$ is defined by the global policy as $\textbf{u}_k^{n,i} = \pi_{\theta_g}(\textbf{u}_{k-1}^{n,i}, \epsilon_c, D_{1:k}^{n,i})$, with $D_k^{n,i} \in \mathbb{R}_+^{N_d \times (3N_p^n + 2N_i^n)}$ containing the oil production rate, water injection rate, watercut and BHPs that satisfy the nonlinear constraints $\textbf{c}(\pi_{\theta_g})$. 
Because the number of wells varies between assets, the input and output of the policy $\pi_{\theta_g}$ varies in size. New treatments are thus required to extend the single-asset control policy representation to the multi-asset setting. 

\subsection{Global control policy representation}

We represent the historical data for each asset $n$ in control step $k$ by the global policy state update function given by 
\begin{equation}
   \hat{\textbf{s}}_{k}^n = f_{\zeta_g}(\hat{S}_{k-1}^n, \textbf{u}_{k-1}^n, \epsilon_c, D_{k}^n),
    \label{eq:ma_mem_upd}
\end{equation}
where the parameters $\boldsymbol{\zeta}_g \subset \boldsymbol{\theta}_g$. The state update function $f_{\zeta_g}$ must be able to handle the varying dimensions of the well settings and observed data. Using a similar treatment as in the single-asset control policy formulation, the previous well settings and current observation for asset $n$ are defined by the matrix $X_k^n \in \mathbb{R}^{N_{d}\times(4N_p^n + 3N_i^n)}_+$. Note that $X_k^n$ does not include the user-specified $\epsilon_c$ for the well settings. We define a unified input to the global control policy as $X_{g,k} = [X_k^1 \ X_k^2 \ \hdots \ X_k^{N_a} \ \boldsymbol{\epsilon}_c]$. This matrix is a row-wise concatenation of the inputs from all assets. Here $X_{g,k} \in \mathbb{R_+}^{N_d \times (4N_{pt} + 3N_{it} + 1)}$, with $N_{pt} = \sum\limits_{n = 1}^{N_a} N_p^n$, $N_{it} = \sum\limits_{n = 1}^{N_a} N_i^n$, and $\boldsymbol{\epsilon}_c = \epsilon_c\textbf{I}$, where $\textbf{I}$ is an $N_d$-dimensional unit vector. 

During training, realizations from each asset are sampled. When a realization from asset $n$ is sampled, entries to $X_k^n$ and $\boldsymbol{\epsilon}_c$ are obtained. Therefore, $X_k^n$ and $\boldsymbol{\epsilon}_c$ provide the input to the global control policy $X_{g,k}$, with the input data for other assets $m \neq n$ set to a null matrix, i.e., $X_k^m = \textbf{0} \ \forall m \neq n$. 

\begin{figure*}[!b]
	\centering
	\includegraphics[width=0.8\textwidth]{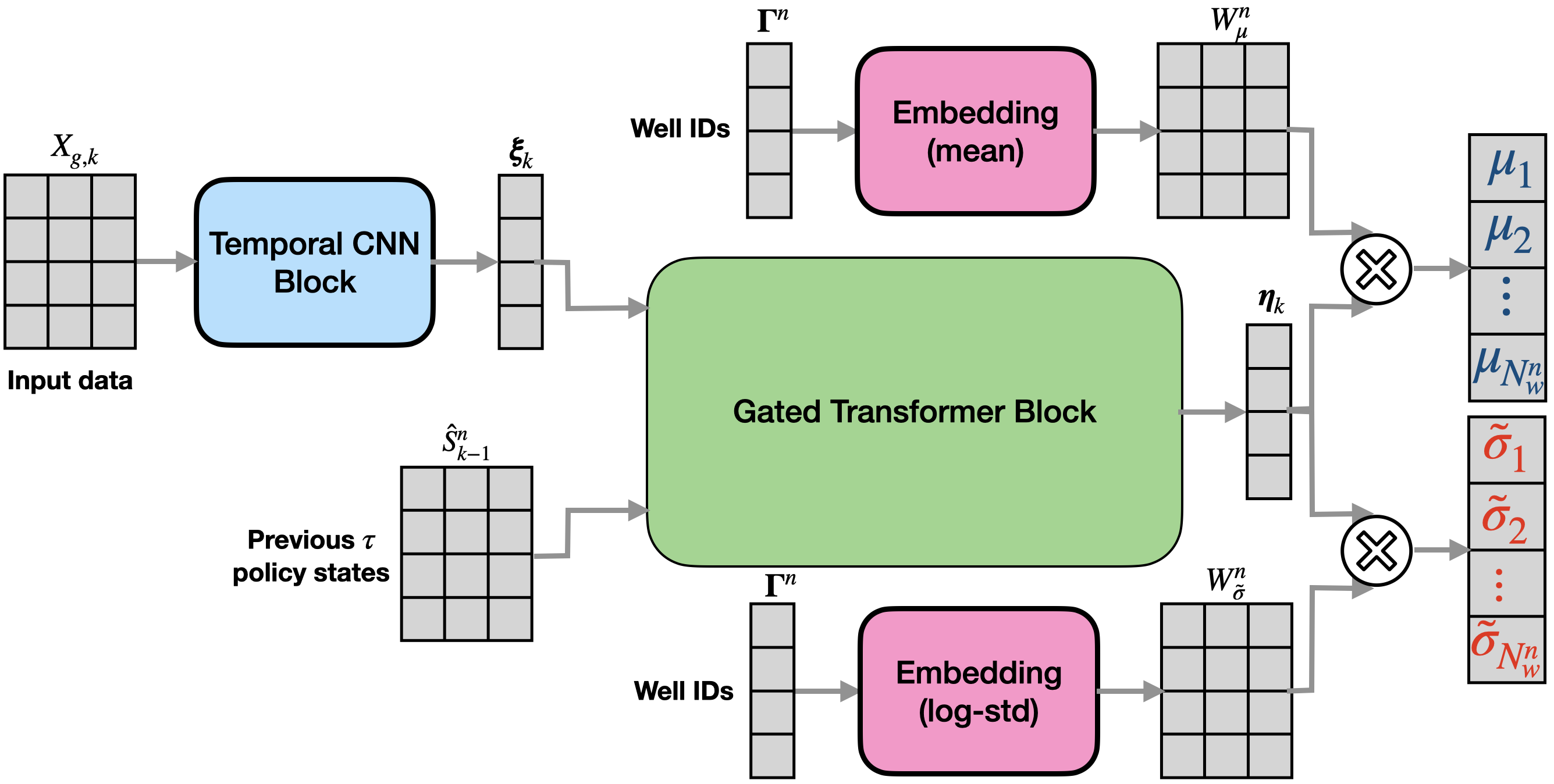}
	\caption{Multi-asset control policy represented by temporal and gated transformer blocks and embedding layers.}
	\label{fig:policy_multi_asset}
\end{figure*}

In the single-asset policy formulation, the output of the gated transformer $\boldsymbol{\eta}_k$ is processed by a fully-connected layer to obtain the mean and log-standard deviation of the action distribution. During the training of the global control policy, the dimensions of the mean and log-standard deviation depend on the number of wells in the sampled asset. To handle the varying dimension of the action distribution, we place an embedding layer between the output of the gated transformer block and the action distribution, as shown in Fig.~\ref{fig:policy_multi_asset}. Embedding layers of this type have been widely applied during the training of neural networks to represent categorical variables as continuous vectors. In natural language processing, for example, the words in a dictionary are represented by word embeddings~\citep{mikolov2013efficient, mikolov2013distributed}, where words with similar meaning are close in the embedding space.

In reinforcement learning, embeddings were used by~\citep{berner2019dota} to represent actions in a strategy game. At each decision-making stage, the embeddings for the available actions were used to define the action distribution. The actions in that case were discrete, so the action distribution was defined by the probability of taking each available action. Our actions however are continuous, and we need to define the Gaussian action distribution. As shown in Fig.~\ref{fig:policy_multi_asset}, we train two embedding layers for the mean and log-standard deviation of the action distribution. The inputs to the embedding layers are the well identifiers (IDs) for each asset. We define the well IDs for asset $n$, $\boldsymbol{\Gamma}^n \in \mathbb{R}_+^{N_w^n}$, as
%
%
\begin{equation} \label{eq:well_id}
\boldsymbol{\Gamma}^n = [N_{wt}^{n-1}+1, \hdots, N_{wt}^{n-1}+N_w^n],
\end{equation}
where $N_{wt}^{n-1} = N_w^1 + N_w^2 + \hdots + N_w^{n-1}$. This means each of the $N_{wt} = \sum_{n = 1}^{N_a} N_w^n$ wells has a unique value in the range $[1, N_{wt}]$. The outputs from the mean and log-standard deviation embedding layers are denoted by $W_{\mu}^n \in \mathbb{R}^{N_w^n \times N_m}$ and $W_{\tilde{\sigma}}^n \in \mathbb{R}^{N_w^n \times N_m}$, respectively, where each row is an embedding for a particular well. The entries in $W_{\mu}^n$ and $W_{\tilde{\sigma}}^n$ are learned
during the global control policy optimization. The action mean $\boldsymbol{\mu}_k^n \in \mathbb{R}^{N_w^n}$ and log-standard deviation $\tilde{\boldsymbol{\sigma}}_k^n \in \mathbb{R}^{N_w^n}$ for asset $n$ in control step $k$ are given by
\begin{subequations}
\begin{align}
\label{eq:ma_act_dist}
\begin{split}
\boldsymbol{\mu}_k^n = W_{\mu}^n\boldsymbol{\eta}_k,
\end{split}
\\[2ex]
\begin{split}
\tilde{\boldsymbol{\sigma}}_k^n = W_{\tilde{\sigma}}^n\boldsymbol{\eta}_k.
\end{split}
\end{align}
\end{subequations}
%

Learning in the global control policy training involves the translation of an `intention,' encoded in $\boldsymbol{\eta}_k$ by the temporal CNN and gated transformer blocks, to the appropriate action for each asset through the embeddings. Such translation is not necessary in the single-asset case as $\boldsymbol{\eta}_k$ is linearly mapped to the action distribution of a single asset. Due to the larger number of prior geological realizations ($N_a \times N_r$) involved in the multi-asset CLRM, the number of parameters in the global control policy should be greater than that in the policies for the individual assets, i.e., $N_g > N^n$, where $N^n$ is the number of parameters in the control policy for any asset $n$. 

The temporal CNN blocks in the control policy architecture (Figs.~\ref{fig:policy_single_asset} and \ref{fig:policy_multi_asset}) contain two 1D CNN layers, each with 64 filters and a filter size of 3. We set the dimension, $N_m$, of the output features for both the temporal CNN block and gated transformer block to 128. Similar to~\citep{nasir2022deep}, we consider a two-layer transformer block. The MLP submodule of the transformer block contains two fully-connected layers. In the single-asset case, the first fully-connected layer in the MLP submodule has 64~units and the second has 128~units. These fully-connected layers both have 128~units in the multi-asset CLRM.


\section{Numerical Results}
\label{sec:results}

We now evaluate the performance of the multi-asset CLRM technique for two example cases. In the first case, assets defined by 2D reservoir models with different numbers of wells but identical areal dimensions are considered. The second example is more general and involves 3D reservoir models of varying dimensions and well counts. In both cases, four distinct assets are considered. We compare the performance of the global control policy trained using the new multi-asset CLRM procedure to that of individual policies trained separately for each asset.

\subsection{Problem setup}

\begin{figure}[htbp!]
	\centering
	\includegraphics[width=8cm]{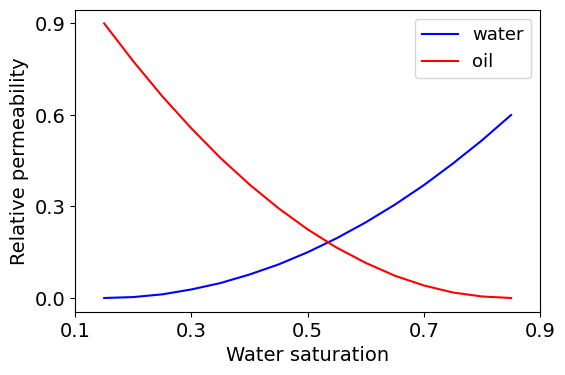}
	\caption{Oil-water relative permeability curves used in both example cases.}
	\label{fig:relperm}
\end{figure}

We consider water-flooding problems with different numbers of injection and production wells in each of the four assets. In both the 2D and 3D cases, the initial reservoir pressure is 350~bar, and the initial oil saturation is 0.85. The oil and water densities at reservoir conditions are 849~kg/m$^3$ and 1025~kg/m$^3$. Oil and water viscosities are set to 1~cp and 0.31~cp. Relative permeability curves for the two phases are shown in Fig.~\ref{fig:relperm}. There is no free gas in the system, and the effects of capillary pressure are neglected. All simulations are performed using the Delft Advanced Research Terra Simulator~\citep{khait2017operator, khait2019delft}.

A total of 1000 prior realizations are generated for each asset in both the 2D and 3D examples. Permeability realizations are constructed using sequential Gaussian simulation~\citep{SGeMs}. The isotropic variogram range (given in terms of the number of grid blocks) and type (exponential or spherical) used for each asset are provided in Table~\ref{tab:var_assets}. The numbers of production and injection wells for each asset are also shown in the table. Porosity (which is constant in all models) is set to 0.2. 

\begin{table*}[h]
  \centering
  \caption{Variogram parameters characterizing the geological models and the number of wells for each of the four assets. These values are the same in both the 2D and 3D cases}
  \label{tab:var_assets}
  \begin{tabular}{c c c c c }
  	\hline 
    Asset & Range  & Type & Num. producers & Num. injectors  \\
      &  (blocks) &  \\
     \Xhline{3\arrayrulewidth}
      A &  20  & exponential & 5   & 4 \\
      B &  25  & exponential & 12  & 4 \\
      C &  30  & spherical   & 8   & 3 \\
      D &  25  & spherical   & 9   & 5 \\
     \hline
  \end{tabular}
\end{table*}

An initial production period of 200~days, in which injection and production wells operate at fixed BHPs of 400~bar and 345~bar, is specified. Well-rate data are collected during this initial period. Production optimization begins at day~200. Each control step extends for 200~days, and the maximum project life is 4000~days. The optimization therefore involves a maximum of 19 control steps. As noted in Section~\ref{sec:cp_clrm}, the project is terminated at any control step that gives a negative NPV contribution. Therefore, the project life of an asset may be less than 4000~days. The allowable range for production well BHPs is between 280 and 345~bar, and that for injection wells is between 355 and 450~bar. Well liquid production and water injection rates are constrained to a maximum of 1526~m$^3$/day, equivalent to about 9600~STB/day (note these are nonlinear output constraints). For the NPV computation, the oil price is \$70/STB, the operating cost is \$41,000/day, the costs of produced and injected water are both \$7/STB, and the discount rate is 0.1.

The number of learnable parameters for the individual control policies are 605,523, 612,705, 609,057, and 610,977 for Asset~A, B, C, and D, respectively. The control policy for the multi-asset CLRM involves 940,033 parameters. Training of the control policies is achieved through stochastic gradient descent with the Adam optimizer~\citep{kingma2014adam}. We perform 10~epochs per policy training iteration, with a batch size of 128. The learning rate is initially set to $10^{-4}$. This rate decreases linearly, to $10^{-5}$, at the final iteration.

During training of the control policies, a subset of realizations is simulated at each iteration. The simulated realizations are sampled from the 1000 realizations of each asset. The realizations selected for simulation are intended to be representative of the flow behavior of the full set. This is accomplished by dividing the 1000 realizations for each asset into 40 representative clusters using the approach proposed in \citep{shirangi2016general}. The most representative (centroid) realization in each cluster is excluded from the policy training. This set of 40 excluded realizations is used for evaluation of the policies after training. 

For the training of individual control policies for each asset, we simulate 160 randomly sampled realizations (four from each cluster). These simulations can be fully parallelized (given access to 160 processors). The training is terminated after 1000 iterations, resulting in a total of 160,000 training simulations for each individual control policy. 

\begin{figure*}[!hb]
    \centering
    \begin{subfigure}[b]{0.32\textwidth}
        \includegraphics[width=1\textwidth]{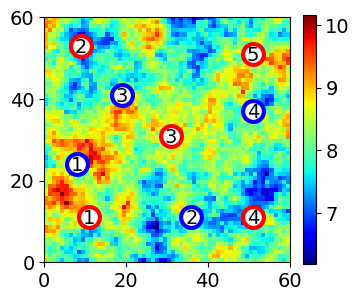}
        \caption{Asset A}
    \end{subfigure}%
    ~
    \begin{subfigure}[b]{0.32\textwidth}
        \includegraphics[width=1\textwidth]{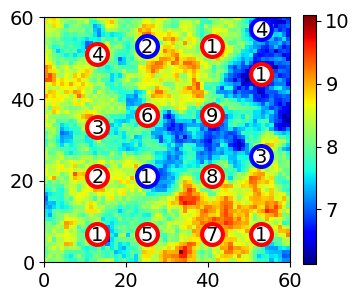}
        \caption{Asset B}
    \end{subfigure}%
    
     \begin{subfigure}[b]{0.32\textwidth}
        \includegraphics[width=1\textwidth]{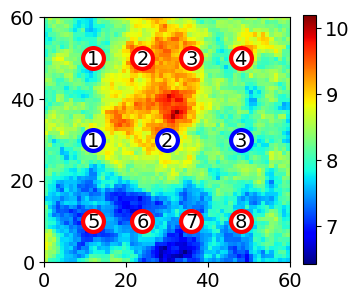}
        \caption{Asset C}
    \end{subfigure}%
    ~
    \begin{subfigure}[b]{0.32\textwidth}
        \includegraphics[width=1\textwidth]{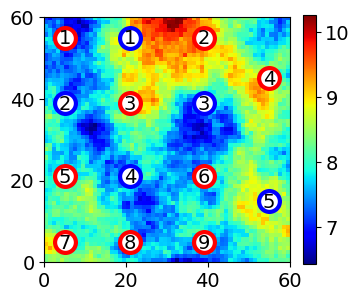}
        \caption{Asset D}
    \end{subfigure}%

    \caption{One realization ($\log_e$ permeability in md is shown) for each of the four assets. Red circles denote producers and blue circles represent injectors (Example~1). 
    }
    \label{fig:ex_1_realz}
\end{figure*}

At each iteration of the global control policy training, we randomly sample five realizations from each of 40 global clusters. These global clusters contain all models from the corresponding clusters for the four assets. They are combined based simply on cluster index.
The global policy training is terminated after 1000 iterations, resulting in a total of 200,000 simulations. Thus the global control policy training requires a factor of $200,000/(4 \times 160,000)=0.3125$ of the training simulations performed in asset-by-asset training. Note that this ratio will decrease with larger numbers of assets, meaning global training will provide increased benefit as more assets are considered.

\subsection{Example 1: CLRM for 2D reservoir models}

The 2D reservoir models contain $60 \times 60$ grid blocks, with each grid block of dimension $\Delta x = \Delta y = 60$~m and $\Delta z = 12$~m. Figure~\ref{fig:ex_1_realz} shows one realization for each asset along with the wells, which clearly differ between assets. The realizations for each asset are conditioned to the permeability values at the well locations.

\subsubsection{Individual control policy results}

\begin{figure*}[htbp!]
    \centering
    \begin{subfigure}{7.1cm} 
        \includegraphics[width=\textwidth]{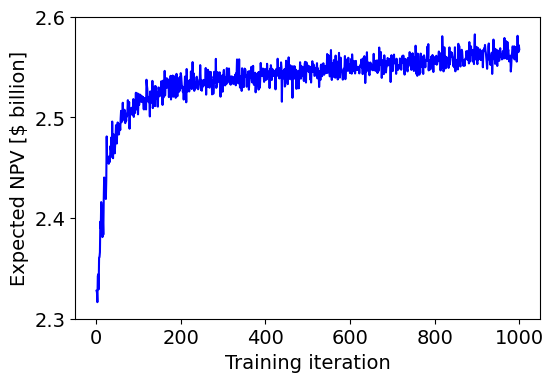}
        \caption{Asset A}
    \end{subfigure}%
    ~
    \begin{subfigure}{7.1cm}
        \includegraphics[width=\textwidth]{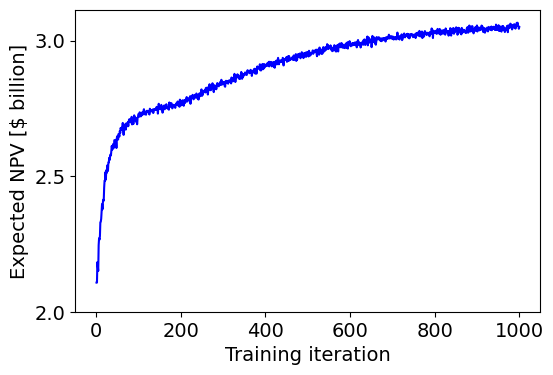}
        \caption{Asset B}
    \end{subfigure}%
    
     \begin{subfigure}{7.1cm}
        \includegraphics[width=\textwidth]{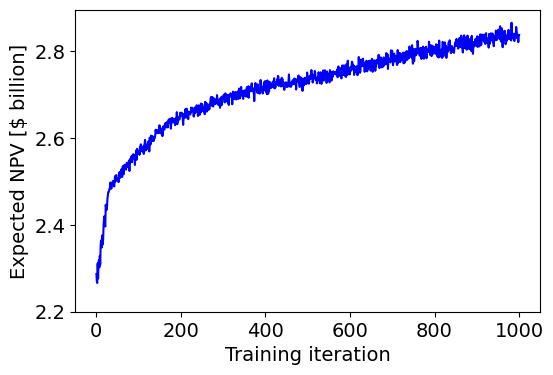}
        \caption{Asset C}
    \end{subfigure}%
    ~
    \begin{subfigure}{7.1cm}
        \includegraphics[width=\textwidth]{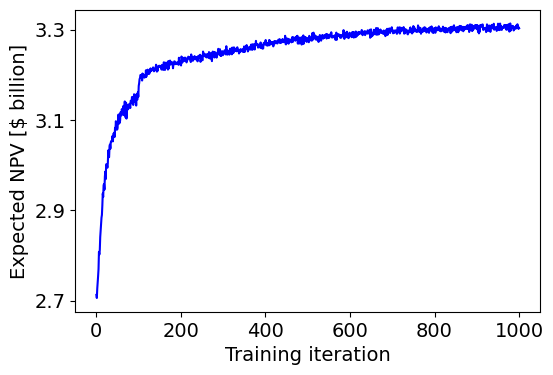}
        \caption{Asset D}
    \end{subfigure}%

    \caption{Evolution of expected NPV during training for each asset (Example~1, individual control policies).}
    \label{fig:training_perf_ex1}
\end{figure*}

We first consider the performance of DRL-based control policies trained separately for each asset. Figure~\ref{fig:training_perf_ex1} displays the evolution of expected NPV for each of the individual control policies. The expected NPV at each iteration is computed with the set of sampled realizations simulated with well settings sampled from the action distribution of the most recent policy. The fluctuations observed in Fig.~\ref{fig:training_perf_ex1} are due to the randomness introduced by the sampling procedures. We observe generally increasing expected NPV over the course of these trainings. Specifically, for Assets~A through D, from the first (randomly initialized policy) iteration to the final iteration, expected NPVs increase by 10.3\%, 44.8\%, 24.1\% and 21.7\%, respectively. 

\begin{figure*}[htbp!]
    \centering
    \begin{subfigure}{7.1cm}
        \includegraphics[width=\textwidth]{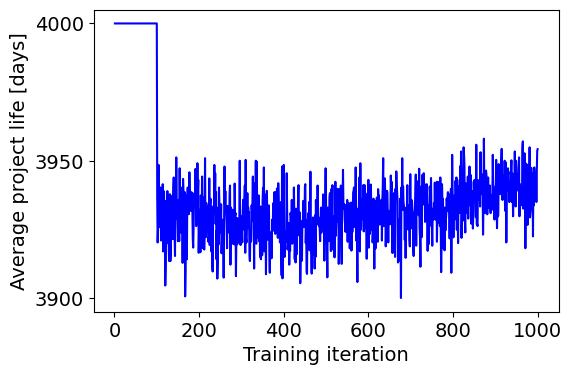}
        \caption{Asset A}
    \end{subfigure}%
    ~
    \begin{subfigure}{7.1cm}
        \includegraphics[width=\textwidth]{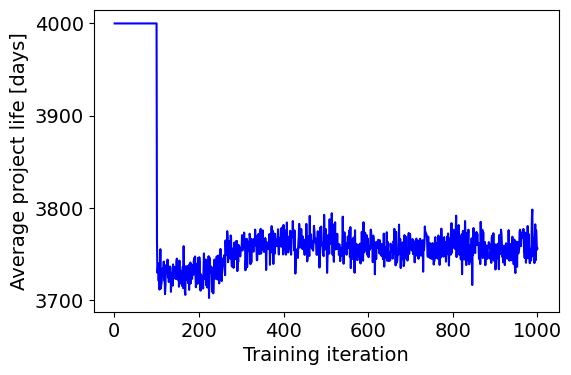}
        \caption{Asset B}
    \end{subfigure}%
    
     \begin{subfigure}{7.1cm}
        \includegraphics[width=\textwidth]{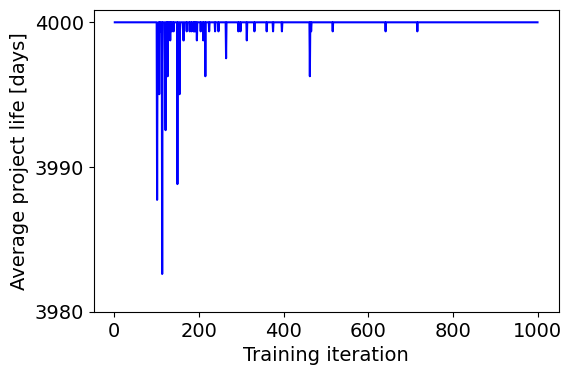}
        \caption{Asset C}
    \end{subfigure}%
    ~
    \begin{subfigure}{7.1cm}
        \includegraphics[width=\textwidth]{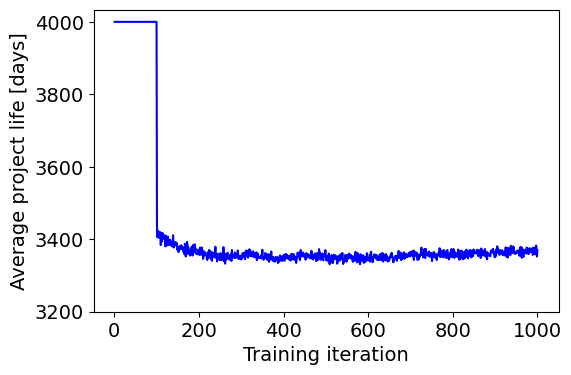}
        \caption{Asset D}
    \end{subfigure}%

    \caption{Evolution of average project life during training for each asset (Example~1, individual control policies).}
    \label{fig:ev_proj_life_ex1}
\end{figure*}

\begin{figure*}[b!]
    \centering
    \begin{subfigure}{7.1cm}
        \includegraphics[width=\textwidth]{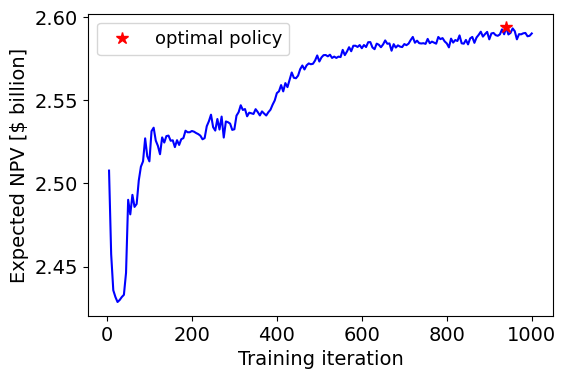}
        \caption{Asset A}
    \end{subfigure}%
    ~
    \begin{subfigure}{7.1cm}
        \includegraphics[width=\textwidth]{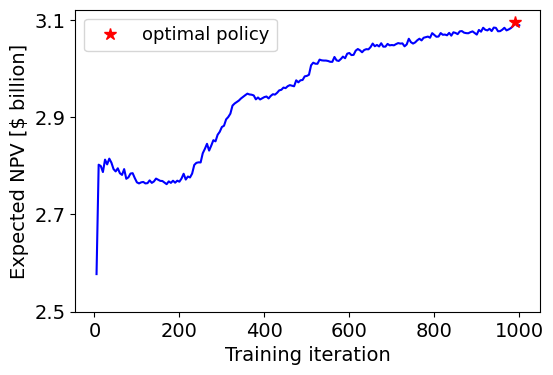}
        \caption{Asset B}
    \end{subfigure}%
    
     \begin{subfigure}{7.1cm}
        \includegraphics[width=\textwidth]{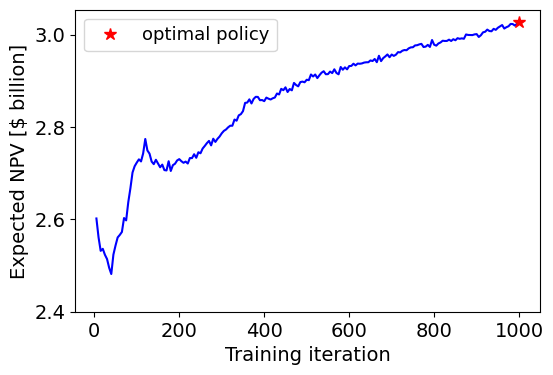}
        \caption{Asset C}
    \end{subfigure}%
    ~
    \begin{subfigure}{7.1cm}
        \includegraphics[width=\textwidth]{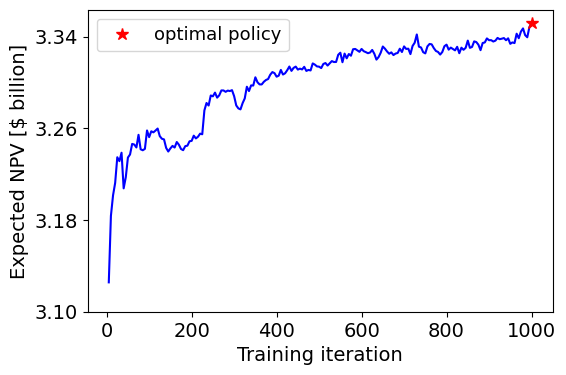}
        \caption{Asset D}
    \end{subfigure}%

    \caption{Evolution of expected NPV for the 40 test-case realizations for each asset with $\epsilon_c = 1$ (Example~1, individual control policies).}
    \label{fig:ev_test_perf_ind_ex1}
\end{figure*}

The evolution of average project life during the training of the individual policies is shown in Fig.~\ref{fig:ev_proj_life_ex1}. We do not allow production to terminate early (before 4000~days) until 100 iterations have been performed. This is because, at early iterations, there is a high degree of randomness in the actions sampled from the action distribution, and this can lead to suboptimal project termination. This premature termination negatively affects the convergence of the policy training. Based on numerical experiments, we found that disallowing termination in the first 100 iterations acts to improve policy convergence. We see from Fig.~\ref{fig:ev_proj_life_ex1} that with this treatment three of the four assets (Asset~C is the exception) still have average project life spans of less than 4000~days. This treatment leads to improved project economics, as expenses associated with daily operations and water handling costs appear in the NPV computation even if little oil is produced.

The optimal control policy from each training run is selected using the test realizations excluded during training. We simulate the 40 test case realizations for each asset, with the well settings defined by the updated policy, after every 10~iterations. The control policy with the highest expected NPV is selected as the optimal policy. The parameter $\epsilon_c$ that defines the allowable relative change for the well settings from control step to control step is set to 1 (unconstrained case) in all simulations performed for these selections. Optimal policy selections based on a range/average of $\epsilon_c$ values could also be considered. As we will see, however, the use of $\epsilon_c=1$ leads to optimal policies that generalize between different $\epsilon_c$ values. Figure~\ref{fig:ev_test_perf_ind_ex1} shows the evolution of the expected NPV computed over the 40 test-case realizations for each asset. The optimal policy for each asset is the maximum achieved at any iteration.

\begin{figure*}[htbp!]
    \centering
    \begin{subfigure}{6cm}
        \includegraphics[width=\textwidth]{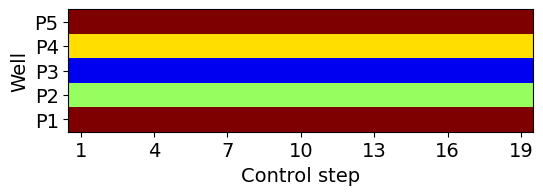}
        \caption{Asset A, $\epsilon_c = 0$}
    \end{subfigure}%
    ~
    \begin{subfigure}{6cm}
        \includegraphics[width=\textwidth]{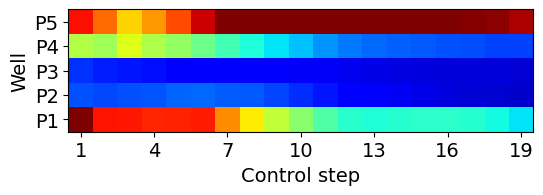}
        \caption{Asset A, $\epsilon_c = 0.5$}
    \end{subfigure}%
    ~
    \begin{subfigure}{6cm}
        \includegraphics[width=\textwidth]{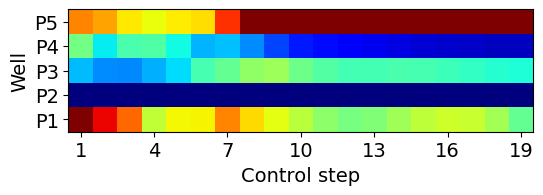}
        \caption{Asset A, $\epsilon_c = 1$}
    \end{subfigure}%
    
    \begin{subfigure}{6cm}
        \includegraphics[width=\textwidth]{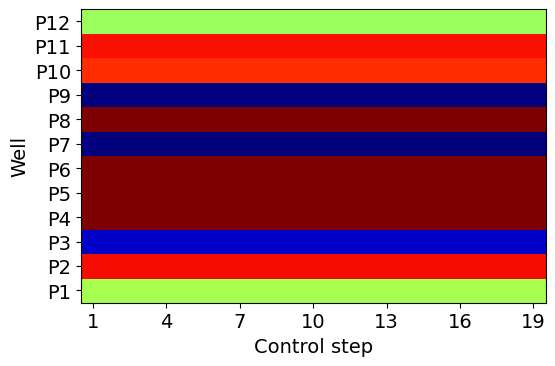}
        \caption{Asset B, $\epsilon_c = 0$}
    \end{subfigure}%
    ~
    \begin{subfigure}{6cm}
        \includegraphics[width=\textwidth]{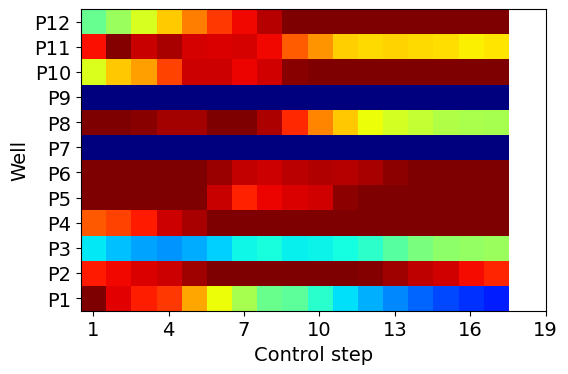}
        \caption{Asset B, $\epsilon_c = 0.5$}
    \end{subfigure}%
    ~
    \begin{subfigure}{6cm}
        \includegraphics[width=\textwidth]{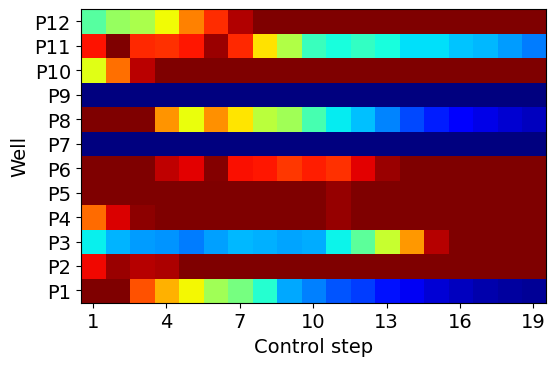}
        \caption{Asset B, $\epsilon_c = 1$}
    \end{subfigure}%
    
    \begin{subfigure}{6cm}
        \includegraphics[width=\textwidth]{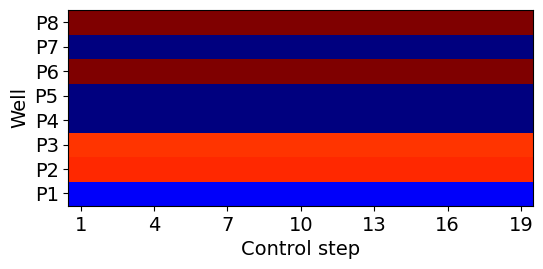}
        \caption{Asset C, $\epsilon_c = 0$}
    \end{subfigure}%
    ~
    \begin{subfigure}{6cm}
        \includegraphics[width=\textwidth]{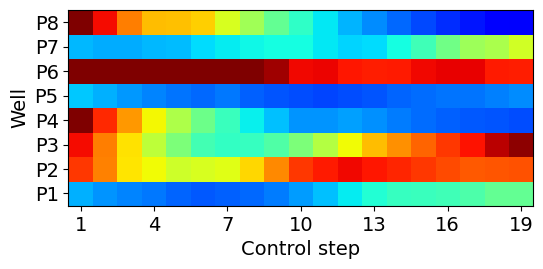}
        \caption{Asset C, $\epsilon_c = 0.5$}
    \end{subfigure}%
    ~
    \begin{subfigure}{6cm}
        \includegraphics[width=\textwidth]{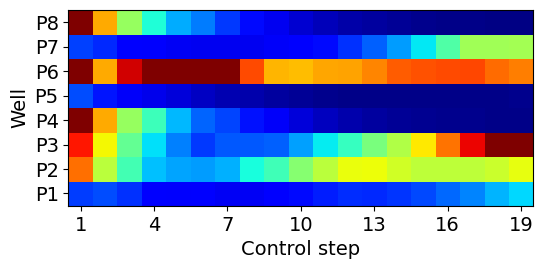}
        \caption{Asset C, $\epsilon_c = 1$}
    \end{subfigure}%
    
    \begin{subfigure}{6cm}
        \includegraphics[width=\textwidth]{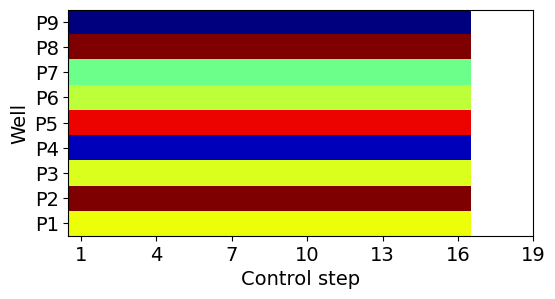}
        \caption{Asset D, $\epsilon_c = 0$}
    \end{subfigure}%
    ~
    \begin{subfigure}{6cm}
        \includegraphics[width=\textwidth]{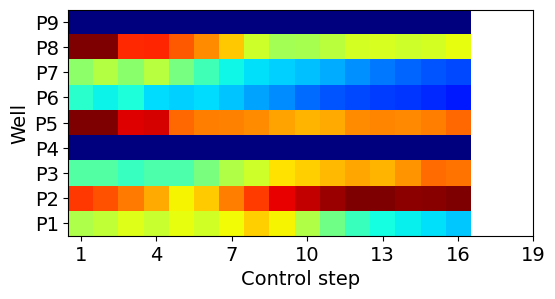}
        \caption{Asset D, $\epsilon_c = 0.5$}
    \end{subfigure}%
    ~
    \begin{subfigure}{6cm}
        \includegraphics[width=\textwidth]{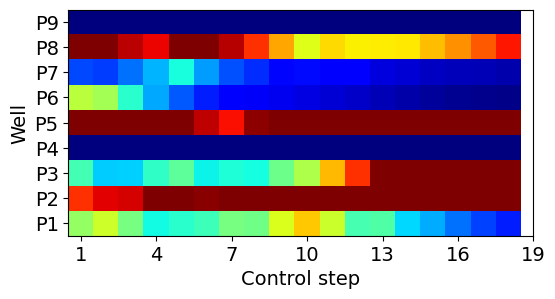}
        \caption{Asset D, $\epsilon_c = 1$}
    \end{subfigure}%
    
    \begin{subfigure}{6cm}
        \includegraphics[width=1\textwidth]{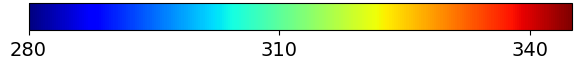}
    \end{subfigure}%

    \caption{Production well BHP settings (in bar) for three different $\epsilon_c$ values. The realization is the same along each row, i.e., for a particular asset. In some cases the production operation terminates before the maximum number of control steps is reached (Example~1, individual control policies).}
    \label{fig:prod_bhp_ex1}
\end{figure*}

Production well BHP settings for particular realizations are shown in Fig.~\ref{fig:prod_bhp_ex1}. The realization is the same for each asset, though the specified value of $\epsilon_c$ changes as we proceed along each row. The specific realization for each asset corresponds to the model that provides the median NPV for $\epsilon_c =1$. Note that, with $\epsilon_c =0$, there is effectively just one control step. There are some similarities in the results for different $\epsilon_c$ values, though generally smoother profiles are observed for $\epsilon_c =0.5$ than for $\epsilon_c =1$, as would be expected. Note also that some wells operate at their minimum BHP regardless of the value of $\epsilon_c$ (e.g., wells P7 and P9 in Asset~B and wells P4 and P9 in Asset~D). 

\begin{figure*}[htbp!]
    \centering
    \begin{subfigure}{7.1cm}
        \includegraphics[width=\textwidth]{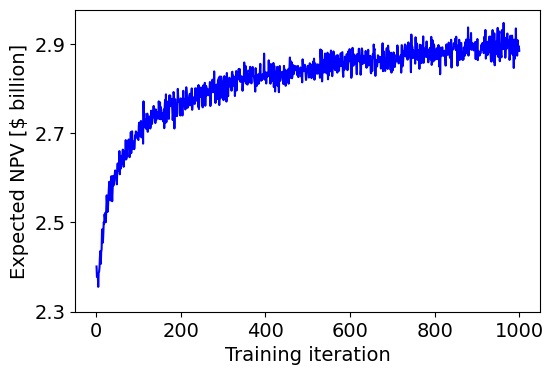}
        \caption{Expected NPV}
        \label{fig:ev_npv_ma_ex1}
    \end{subfigure}%
    ~
    \begin{subfigure}{7.1cm}
        \includegraphics[width=\textwidth]{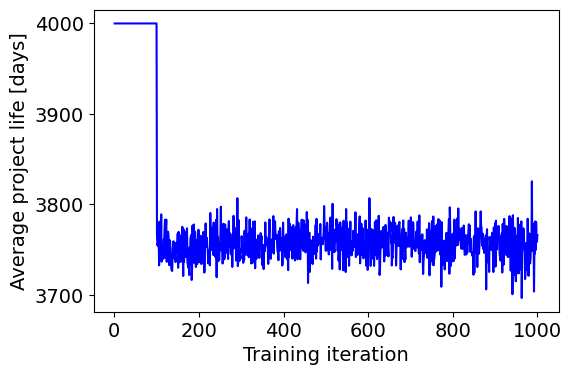}
        \caption{Average project life}
        \label{fig:ev_pl_ma_ex1}
    \end{subfigure}%

    \caption{Evolution of expected NPV and average project life during training (Example~1, global control policy).}
    \label{fig:train_perf_ma_ex1}
\end{figure*}

\begin{figure*}[htbp!]
    \centering
    \begin{subfigure}{7cm}
        \includegraphics[width=\textwidth]{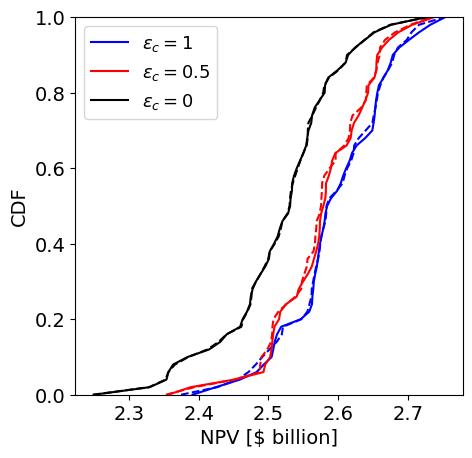}
        \caption{Asset A}
    \end{subfigure}%
    ~
    \begin{subfigure}{7cm}
        \includegraphics[width=\textwidth]{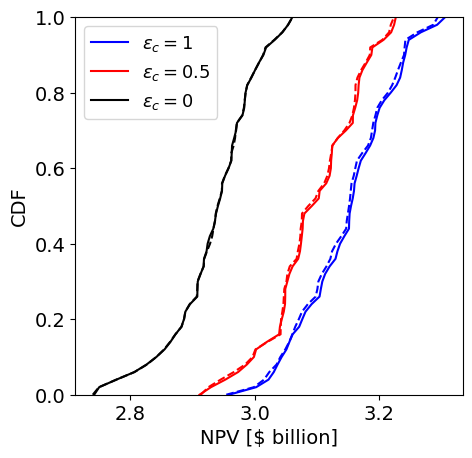}
        \caption{Asset B}
    \end{subfigure}%
    
     \begin{subfigure}{7cm}
        \includegraphics[width=\textwidth]{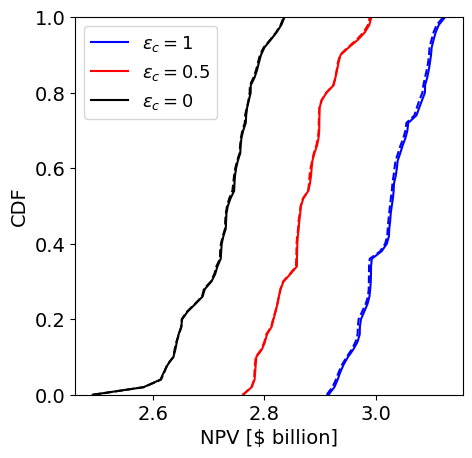}
        \caption{Asset C}
    \end{subfigure}%
    ~
    \begin{subfigure}{7cm}
        \includegraphics[width=\textwidth]{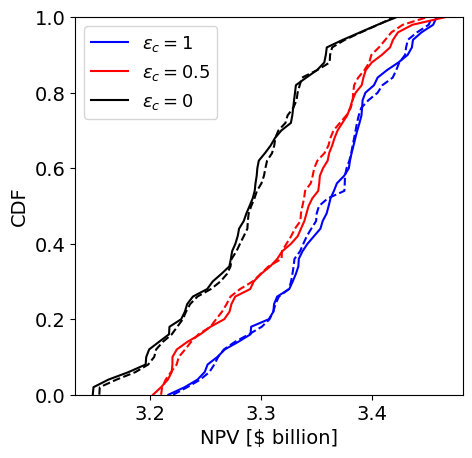}
        \caption{Asset D}
    \end{subfigure}%

    \caption{Comparison of the CDFs for the optimum NPVs of the 40 test-case realizations using the global control policy (dashed curves) and individual control policies (solid curves). In many places the curves overlap, so only one curve is visible (Example~1).}
    \label{fig:sa_vs_ma_ex1}
\end{figure*}

\subsubsection{Global control policy results}

Results until this point have involved the use of control policies trained separately for each asset. We now present results using global control policy training. The evolution of expected NPV during global training is shown in Fig.~\ref{fig:ev_npv_ma_ex1}. Starting from a random policy, an increase of 20.1\% in expected NPV is achieved after 1000 control policy optimization iterations. Figure~\ref{fig:ev_pl_ma_ex1} shows the evolution of the average project life as the training progresses, and we can see that it converges to approximately 3750~days. This project life is close to the average from the individual control policy trainings, shown in Fig.~\ref{fig:ev_proj_life_ex1}.

The policies from the global control policy training run are evaluated by simulating the 160 representative realizations (40 from each asset) every 10 iterations. We set $\epsilon_c = 1$ in these evaluations. The optimal global policy is taken to be the policy that provides the maximum average NPV over these 160 realizations. 

\begin{figure*}[b!]
    \centering
    \begin{subfigure}{6cm}
        \includegraphics[width=\textwidth]{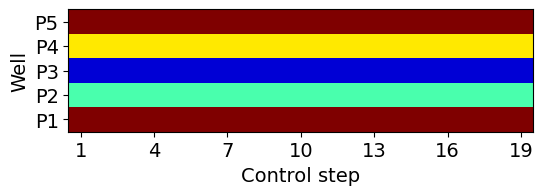}
        \caption{Asset A, $\epsilon_c = 0$}
    \end{subfigure}%
    ~
    \begin{subfigure}{6cm}
        \includegraphics[width=\textwidth]{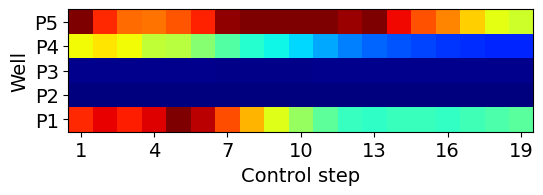}
        \caption{Asset A, $\epsilon_c = 0.5$}
    \end{subfigure}%
    ~
    \begin{subfigure}{6cm}
        \includegraphics[width=\textwidth]{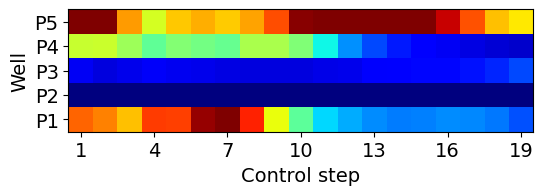}
        \caption{Asset A, $\epsilon_c = 1$}
    \end{subfigure}%
    
    \begin{subfigure}{6cm}
        \includegraphics[width=\textwidth]{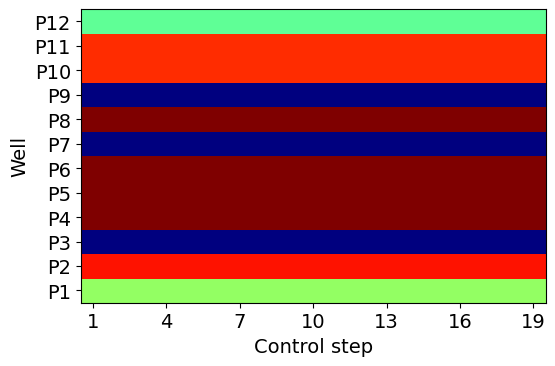}
        \caption{Asset B, $\epsilon_c = 0$}
    \end{subfigure}%
    ~
    \begin{subfigure}{6cm}
        \includegraphics[width=\textwidth]{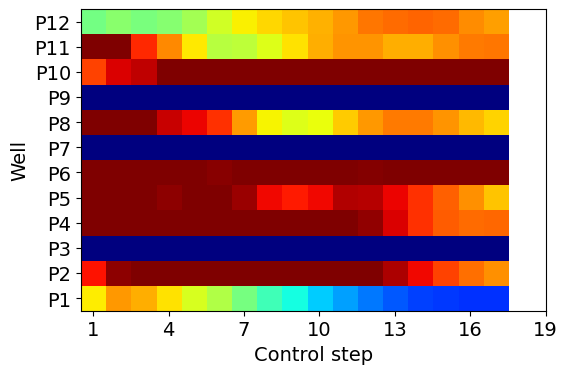}
        \caption{Asset B, $\epsilon_c = 0.5$}
    \end{subfigure}%
    ~
    \begin{subfigure}{6cm}
        \includegraphics[width=\textwidth]{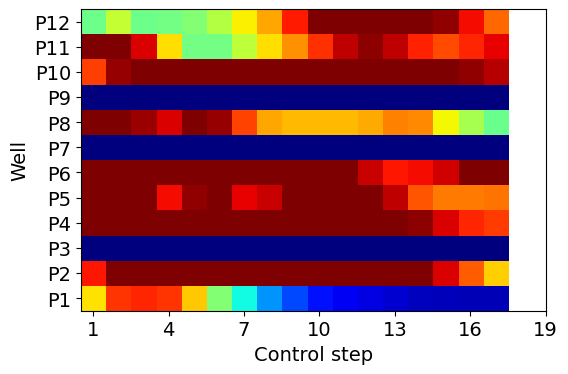}
        \caption{Asset B, $\epsilon_c = 1$}
    \end{subfigure}%
    
    \begin{subfigure}{6cm}
        \includegraphics[width=\textwidth]{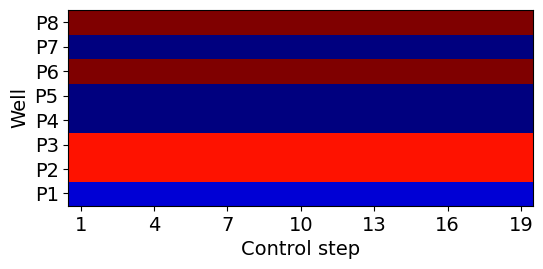}
        \caption{Asset C, $\epsilon_c = 0$}
    \end{subfigure}%
    ~
    \begin{subfigure}{6cm}
        \includegraphics[width=\textwidth]{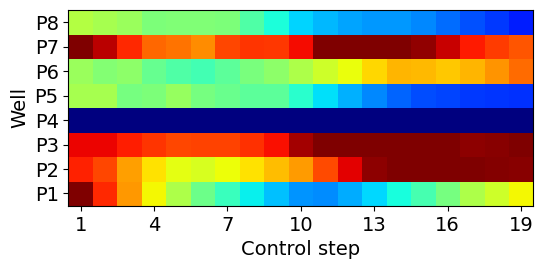}
        \caption{Asset C, $\epsilon_c = 0.5$}
    \end{subfigure}%
    ~
    \begin{subfigure}{6cm}
        \includegraphics[width=\textwidth]{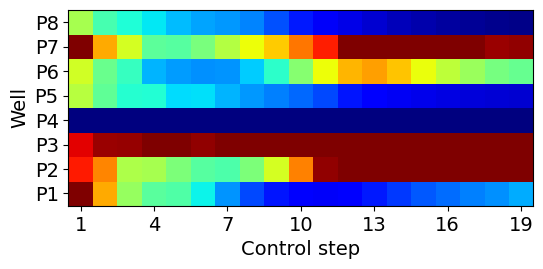}
        \caption{Asset C, $\epsilon_c = 1$}
    \end{subfigure}%
    
    \begin{subfigure}{6cm}
        \includegraphics[width=\textwidth]{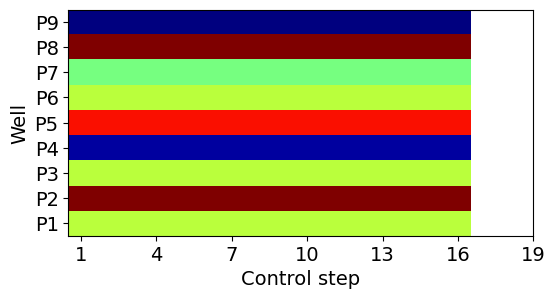}
        \caption{Asset D, $\epsilon_c = 0$}
    \end{subfigure}%
    ~
    \begin{subfigure}{6cm}
        \includegraphics[width=\textwidth]{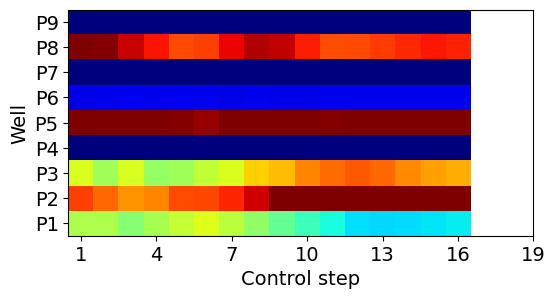}
        \caption{Asset D, $\epsilon_c = 0.5$}
    \end{subfigure}%
    ~
    \begin{subfigure}{6cm}
        \includegraphics[width=\textwidth]{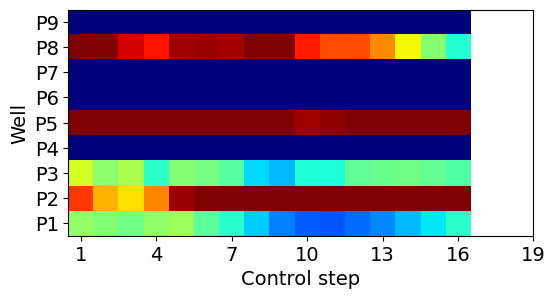}
        \caption{Asset D, $\epsilon_c = 1$}
    \end{subfigure}%
    
    \begin{subfigure}{6cm}
        \includegraphics[width=1\textwidth]{prod_bhp_color_bar.png}
    \end{subfigure}%

    \caption{Production well BHP settings (in bar) for three different $\epsilon_c$ values using the global control policy. The realization considered for each asset is the same as that in Fig.~\ref{fig:prod_bhp_ex1}. Early termination occurs in some cases (Example~1, global control policy).}
    \label{fig:prod_bhp_ma_ex1}
\end{figure*}

We now compare the performance of the global control policy to that of the individual policies for three different values of $\epsilon_c$. Results, presented in terms of cumulative distribution functions (CDFs) for the optimum NPVs over the 40 test-case realizations, are provided for each asset in Figure~\ref{fig:sa_vs_ma_ex1}. In the figure, the solid curves depict results using individual control policies and the dashed curves show results using the global control policy. In many cases these two sets of results essentially overlap, and in all cases there is very close correspondence between them. Additionally, we see that there is generally an increase in NPV as $\epsilon_c$ increases, with the unconstrained case ($\epsilon_c = 1$) displaying the highest NPV. Interestingly, the shifts in the CDFs with $\epsilon_c$ vary from asset to asset.

Production well BHP settings obtained by following the global control policy, for three different $\epsilon_c$ values, are shown in Fig.~\ref{fig:prod_bhp_ma_ex1}. The realization considered for each asset is the same as that in Fig.~\ref{fig:prod_bhp_ex1}. For $\epsilon_c = 0$, the individual control policies and the global control policy provide similar BHPs. For other values of $\epsilon_c$ there is some correspondence between the two sets of BHP profiles, though they differ in detail. For example, in Asset~D we observe early termination (before control step~19) for all values of $\epsilon_c$ in both sets of results, though the precise time of termination differs between Figs.~\ref{fig:prod_bhp_ex1} and \ref{fig:prod_bhp_ma_ex1}.



As noted earlier, in terms of computational cost, the global control policy training entails about 31\% of the flow simulations required to train separate control policies for each asset. The time required for these flow simulations is much larger than that for network training for the cases considered here. In settings with, e.g., many more wells, a larger neural network will be required. In such cases, further computational savings could be realized through use of the multi-asset CLRM procedure, as this requires the training of only a single network. 

\subsection{Example 2: CLRM for 3D reservoir models}

\begin{figure}[htbp!]
    \centering
    \begin{subfigure}[b]{0.34\textwidth}
        \includegraphics[width=1\textwidth]{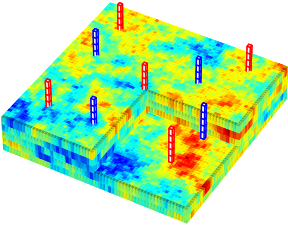}
        \caption{Asset A}
    \end{subfigure}%
    
    \begin{subfigure}[b]{0.34\textwidth}
        \includegraphics[width=1\textwidth]{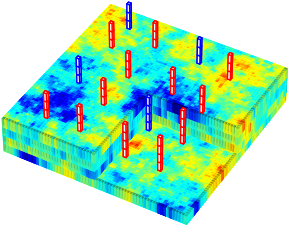}
        \caption{Asset B}
    \end{subfigure}%
    
     \begin{subfigure}[b]{0.34\textwidth}
        \includegraphics[width=1\textwidth]{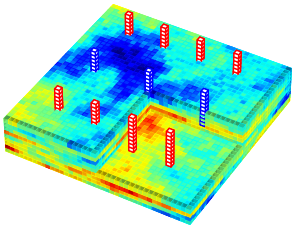}
        \caption{Asset C}
    \end{subfigure}%
    
    \begin{subfigure}[b]{0.34\textwidth}
        \includegraphics[width=1\textwidth]{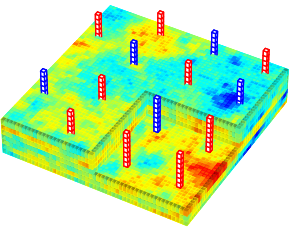}
        \caption{Asset D}
    \end{subfigure}%
    
    \begin{subfigure}[b]{0.3\textwidth}
        \includegraphics[width=1\textwidth]{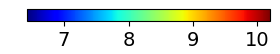}
    \end{subfigure}%

    \caption{One realization ($\log_e$ permeability in md is shown) for each of the four assets. Red pillars denote producers and blue pillars represent injectors (Example~2). Cutaway view shown for 3D visualization.}
    \label{fig:ex_2_realz}
\end{figure}

This example involves 3D reservoir models with varying numbers of areal ($x$-$y$) grid blocks and layers. Model dimensions for the different assets are given in Table~\ref{tab:dim_assets}. In all cases, grid blocks are of size $\Delta x = \Delta y = 60$~m and $\Delta z = 3$~m. Due to the variation in dimensions of the reservoir models, the four assets have different volumes of original oil in place (OOIP), in contrast to the assets considered in Example~1. One realization of the permeability field for each asset is displayed in Fig.~\ref{fig:ex_2_realz}. The well locations, which are similar to those in Example~1, are also shown. Wells are perforated in all layers of each model. Realizations for each asset are again conditioned to permeability values in well blocks. In all cases, the variogram range in the $z$-direction is $3\Delta z$, and the ratio of vertical to horizontal permeability is 0.1.

\begin{table}[h]
  \centering
  \caption{Areal dimension and number of layers for the 3D reservoir models in Example~2}
  \label{tab:dim_assets}
  \begin{tabular}{c c c}
  	\hline 
    Asset & Areal dimension & Num. layers \\
     \Xhline{3\arrayrulewidth}
      A &  60 $\times$ 60 & 5  \\
      B &  65 $\times$ 65 & 4  \\
      C &  40 $\times$ 40 & 9  \\
      D &  50 $\times$ 50 & 7  \\
     \hline
  \end{tabular}
\end{table}

\begin{figure*}[htbp!]
    \centering
    \begin{subfigure}{7.1cm} 
        \includegraphics[width=\textwidth]{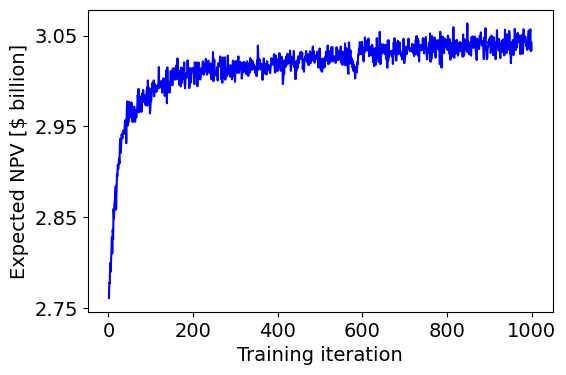}
        \caption{Asset A}
    \end{subfigure}%
    ~
    \begin{subfigure}{7.1cm}
        \includegraphics[width=\textwidth]{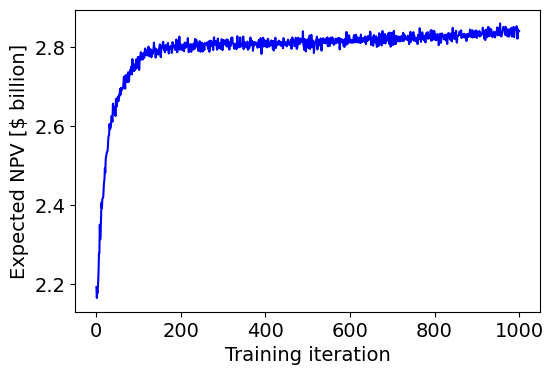}
        \caption{Asset B}
    \end{subfigure}%
    
     \begin{subfigure}{7.1cm}
        \includegraphics[width=\textwidth]{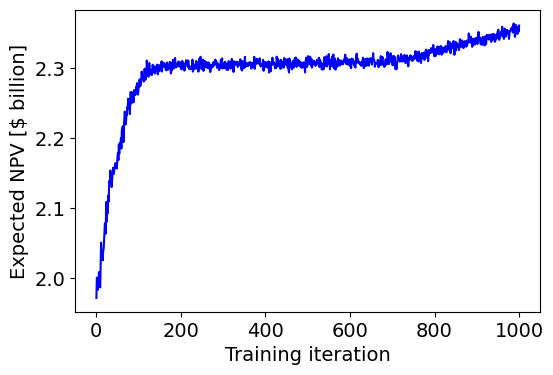}
        \caption{Asset C}
    \end{subfigure}%
    ~
    \begin{subfigure}{7.1cm}
        \includegraphics[width=\textwidth]{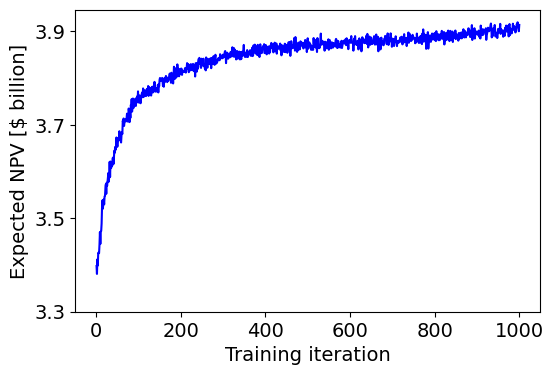}
        \caption{Asset D}
    \end{subfigure}%

    \caption{Evolution of expected NPV during training for each asset (Example~2, individual control policies).}
    \label{fig:training_perf_ex2}
\end{figure*}

The evolution of expected NPV during individual control policy training, for each of the four assets, is shown in Fig.~\ref{fig:training_perf_ex2}. The expected NPV increases by 9.9\%, 29.7\%, 19.8\% and 15.2\% for Assets~A through D, respectively, relative to the (random) initial control policy. Compared to Example~1 (Fig.~\ref{fig:training_perf_ex1}), in this case we see larger variation in expected NPV from asset to asset. This is due in part to the varying OOIP across assets in this example. The average project life (not shown) for the four assets in this example is 4000~days, which is the maximum allowable time frame. 

\begin{figure*}[htbp!]
    \centering
    \begin{subfigure}{7.1cm}
        \includegraphics[width=\textwidth]{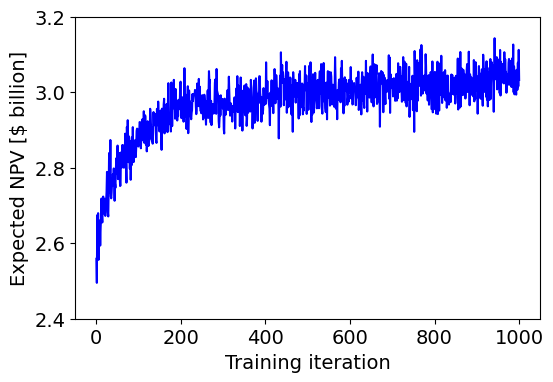}
        \caption{Expected NPV}
        \label{fig:ev_npv_ma_ex2}
    \end{subfigure}%
    ~
    \begin{subfigure}{7.1cm}
        \includegraphics[width=\textwidth]{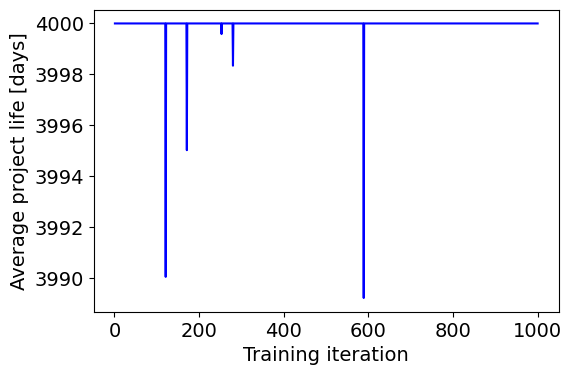}
        \caption{Average project life}
        \label{fig:ev_pl_ma_ex2}
    \end{subfigure}%

    \caption{Evolution of expected NPV and average project life during training (Example~2, global control policy).}
    \label{fig:train_perf_ma_ex2}
\end{figure*}

\begin{figure*}[htbp!]
    \centering
    \begin{subfigure}{7cm}
        \includegraphics[width=\textwidth]{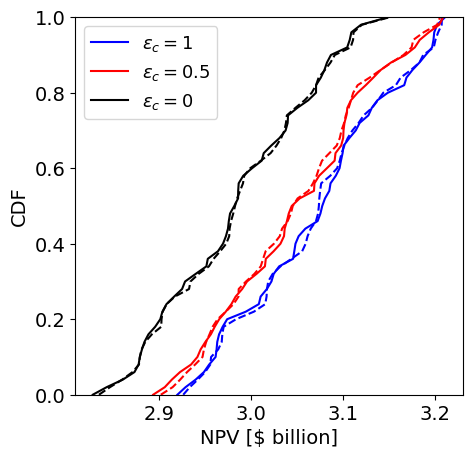}
        \caption{Asset A}
    \end{subfigure}%
    ~
    \begin{subfigure}{7cm}
        \includegraphics[width=\textwidth]{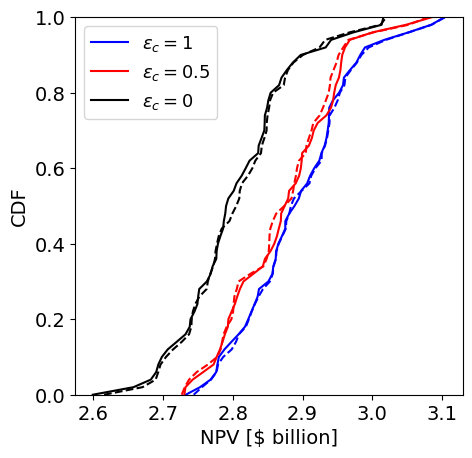}
        \caption{Asset B}
    \end{subfigure}%
    
     \begin{subfigure}{7cm}
        \includegraphics[width=\textwidth]{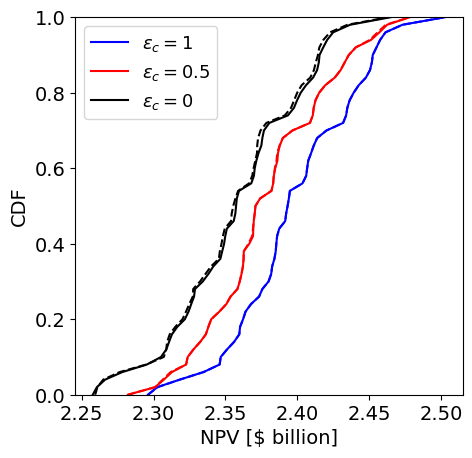}
        \caption{Asset C}
    \end{subfigure}%
    ~
    \begin{subfigure}{7cm}
        \includegraphics[width=\textwidth]{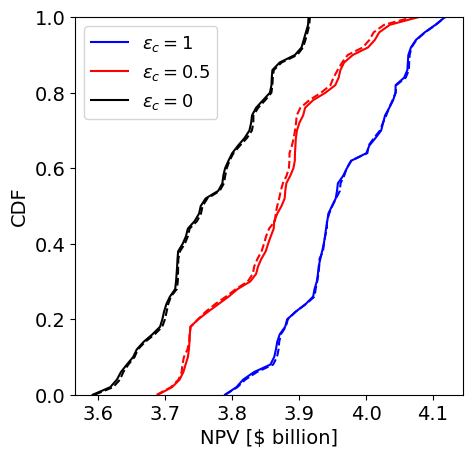}
        \caption{Asset D}
    \end{subfigure}%

    \caption{Comparison of the CDFs for the optimum NPVs of the 40 test-case realizations using the global control policy (dashed curves) and individual control policies (solid curves). In many places the curves overlap, so only one curve is visible (Example~2).}
    \label{fig:sa_vs_ma_ex2}
\end{figure*}

The evolution of expected NPV during global control policy training of the 3D reservoir models is shown in Fig.~\ref{fig:ev_npv_ma_ex2}. Policy training leads to an increase in expected NPV of 18.5\% relative to the random initial policy. The fluctuations in expected NPV for the global control policy in this case are larger than those in Example~1 (Fig.~\ref{fig:ev_npv_ma_ex1}), presumably due to the larger variation between assets. The evolution of the average project life during global policy training is shown in Fig~\ref{fig:ev_pl_ma_ex2}. The average project life converges to 4000~days, which is consistent with the results for the individual control policies.

The optimal individual and global control policies are selected, based on the 40 test-case realizations for each asset, using the same procedure as in Example~1. Figure~\ref{fig:sa_vs_ma_ex2} displays comparisons of the CDFs of the optimum NPVs obtained from the individual and global control policies, for three different $\epsilon_c$ values. We again see very close correspondence between the CDFs, for all assets and all $\epsilon_c$ values. These results, consistent with those obtained in Example~1 (Fig.~\ref{fig:sa_vs_ma_ex1}), highlight the ability of the global control policy to provide optimal settings for multiple assets.

\begin{figure*}[htbp!]
    \centering
    \begin{subfigure}{6cm}
        \includegraphics[width=\textwidth]{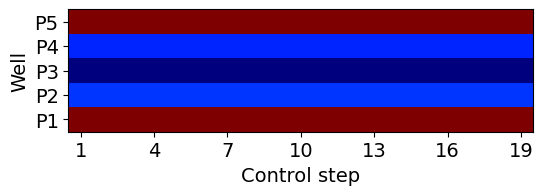}
        \caption{Asset A, $\epsilon_c = 0$}
    \end{subfigure}%
    ~
    \begin{subfigure}{6cm}
        \includegraphics[width=\textwidth]{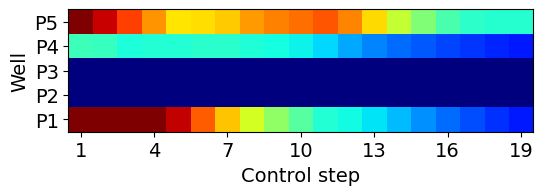}
        \caption{Asset A, $\epsilon_c = 0.5$}
    \end{subfigure}%
    ~
    \begin{subfigure}{6cm}
        \includegraphics[width=\textwidth]{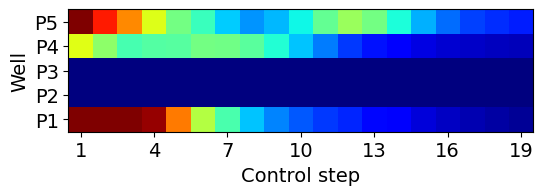}
        \caption{Asset A, $\epsilon_c = 1$}
    \end{subfigure}%
    
    \begin{subfigure}{6cm}
        \includegraphics[width=\textwidth]{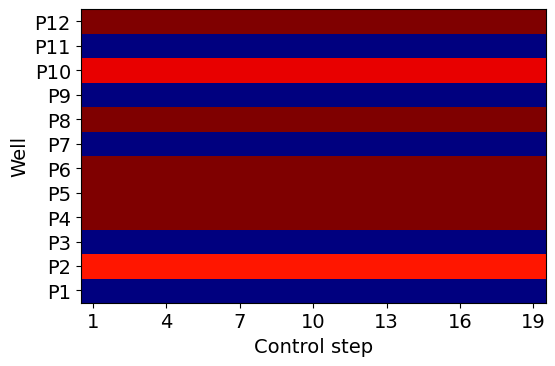}
        \caption{Asset B, $\epsilon_c = 0$}
    \end{subfigure}%
    ~
    \begin{subfigure}{6cm}
        \includegraphics[width=\textwidth]{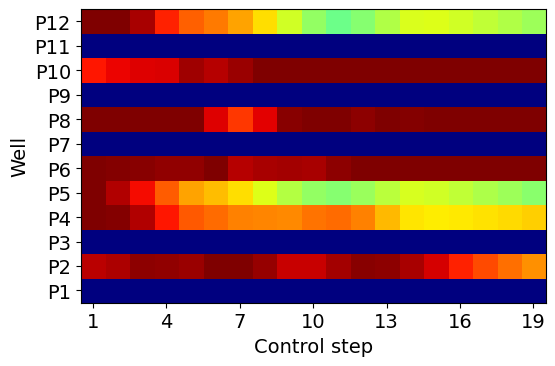}
        \caption{Asset B, $\epsilon_c = 0.5$}
    \end{subfigure}%
    ~
    \begin{subfigure}{6cm}
        \includegraphics[width=\textwidth]{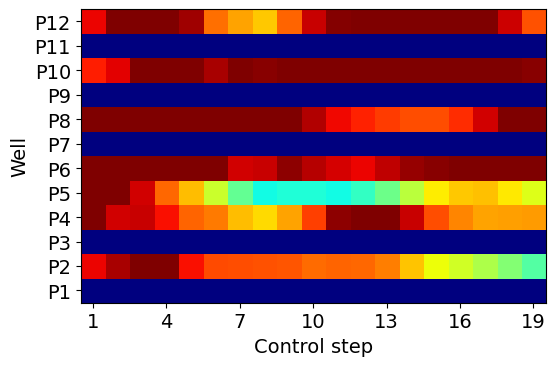}
        \caption{Asset B, $\epsilon_c = 1$}
    \end{subfigure}%
    
    \begin{subfigure}{6cm}
        \includegraphics[width=\textwidth]{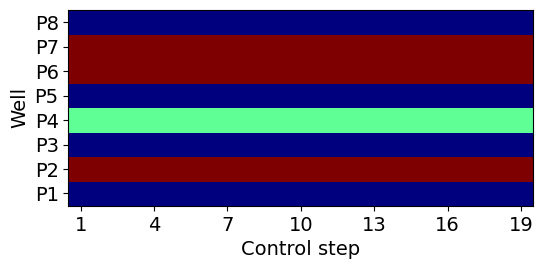}
        \caption{Asset C, $\epsilon_c = 0$}
    \end{subfigure}%
    ~
    \begin{subfigure}{6cm}
        \includegraphics[width=\textwidth]{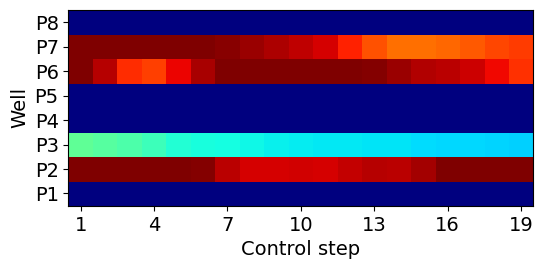}
        \caption{Asset C, $\epsilon_c = 0.5$}
    \end{subfigure}%
    ~
    \begin{subfigure}{6cm}
        \includegraphics[width=\textwidth]{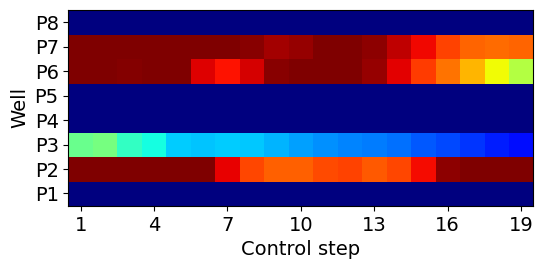}
        \caption{Asset C, $\epsilon_c = 1$}
    \end{subfigure}%
    
    \begin{subfigure}{6cm}
        \includegraphics[width=\textwidth]{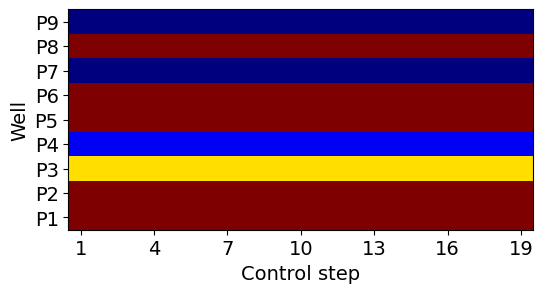}
        \caption{Asset D, $\epsilon_c = 0$}
    \end{subfigure}%
    ~
    \begin{subfigure}{6cm}
        \includegraphics[width=\textwidth]{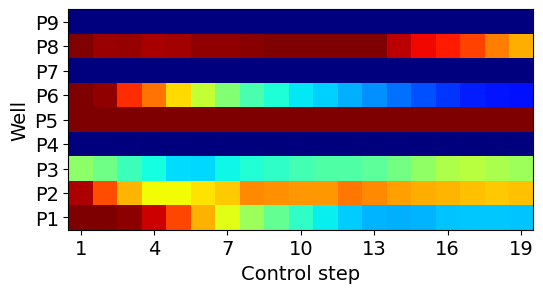}
        \caption{Asset D, $\epsilon_c = 0.5$}
    \end{subfigure}%
    ~
    \begin{subfigure}{6cm}
        \includegraphics[width=\textwidth]{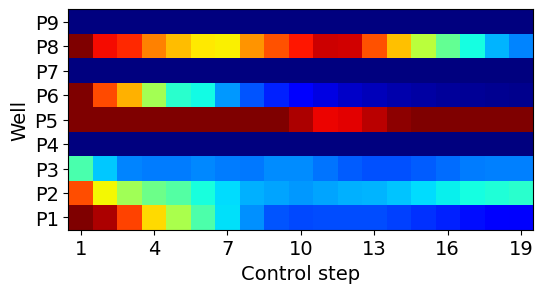}
        \caption{Asset D, $\epsilon_c = 1$}
    \end{subfigure}%
    
    \begin{subfigure}{6cm}
        \includegraphics[width=1\textwidth]{prod_bhp_color_bar.png}
    \end{subfigure}%

    \caption{Production well BHP settings (in bar) for three different $\epsilon_c$ values. The realization is the same along each row, i.e., for a particular asset (Example~2, global control policy).}
    \label{fig:prod_bhp_ma_ex2}
\end{figure*}

Production well BHPs obtained from the global control policy, for three different $\epsilon_c$ values, are shown in Fig.~\ref{fig:prod_bhp_ma_ex2}. The model with the median NPV for $\epsilon_c = 1$ is considered for each asset. Consistent with the results in Example~1 (Fig.~\ref{fig:prod_bhp_ex1}), there is correspondence in the BHP profiles. This is particularly evident in the $\epsilon_c=0.5$ and $\epsilon_c=1$ cases, where we see the same general trends, though with increased variability for $\epsilon_c=1$.

\FloatBarrier

\clearpage
\section{Concluding Remarks}
\label{sec:conclusion}
In this work, we developed a control policy framework based on deep reinforcement learning for closed-loop reservoir management (CLRM) over multiple assets. The framework enables the training of a single global control policy that maps observed well quantities to optimal well pressure settings. The formulation is applicable for different assets with varying well counts and configurations. In contrast to the single-asset CLRM~\citep{nasir2022deep}, the representation of the global control policy includes embedding layers to accommodate the different numbers of decision variables in the multi-asset setting. We also incorporated a relative-change constraint in our formulation to avoid large jumps in well settings from one control step to another, consistent with practical considerations.

The new framework was tested on two multi-asset problems involving 2D and 3D models. In both cases, we considered four assets containing different numbers of wells and characterized by different geostatistical parameters. In each example case, the performance of the global control policy was compared to that of control policies trained individually for each asset. We observed close agreement between the optimum NPVs obtained from the global and individual control policies for all assets and values of the relative-change parameter, $\epsilon_c$. Training of the global control policy required only about 31\% of the computational cost expended in training the individual control policies, and this speedup is expected to increase if more assets are considered. Cost reduction is due to the ability of the global control policy to learn useful features across different assets. This is in contrast to the training of individual control policies, where informative features must be learned independently.

There are a number of research directions that could be explored in future work. Further reduction in the computational cost of the global control policy training could be achieved through use of deep learning based surrogates~\citep{kim2022recurrent, wang2022deep}. Extension of these proxies to multi-asset settings will enhance their utility. Application of the framework to other recovery processes and CO$_2$ storage operations should also be investigated. In the latter case, the objective would be to minimize a measure of risk or cost. Finally, the framework should be tested on field cases of varying complexity.

\begin{acknowledgements}
We are grateful to the Stanford Center for Computational Earth \& Environmental Sciences for providing computational resources. 
\end{acknowledgements}

\small {\noindent {\bf Funding information} The authors received financial support from the Stanford Graduate Fellowship program and the industrial affiliates of the Stanford Smart Fields Consortium.}

\small {\noindent {\bf Competing interests} We declare that this research was conducted in the absence of any commercial or financial relationships that could be construed as a potential conflict of interest.}

\small {\noindent {\bf Data availability} Inquiries regarding the data should be directed to Yusuf Nasir, nyusuf@stanford.edu.}

\bibliographystyle{elsarticle-num-names} 
\bibliography{ref}

\end{document}